# Large linguistic models: Investigating LLMs' metalinguistic abilities

Gašper Beguš, *Member, IEEE*, Maksymilian Dabkowski, and Ryan Rhodes

***Abstract*—The performance of large language models (LLMs) has recently improved to the point where models can perform well on many language tasks. We show here that—for the first time—the models can also generate valid metalinguistic analyses of language data. We outline a research program where the *behavioral interpretability* of LLMs on these tasks is tested via prompting. LLMs are trained primarily on text—as such, evaluating their metalinguistic abilities improves our understanding of their general capabilities and sheds new light on theoretical models in linguistics. We show that OpenAI's (2024) o1 vastly outperforms other models on tasks involving drawing syntactic trees and phonological generalization. We speculate that OpenAI o1's unique advantage over other models may result from the model's *chain-of-thought* mechanism, which mimics the structure of human reasoning used in complex cognitive tasks, such as linguistic analysis.**

***Impact Statement*—The nature of large language models' (LLMs) internal linguistic representations poses a consequential question in generative AI research. We outline a research program where we test the *behavioral interpretability* of LLMs such as OpenAI's (2024) o1 by prompting them to conduct linguistic analyses. We argue that these tasks present the perfect testing ground for accessing the LLMs' higher-level "cognitive" abilities, inform us about their general capabilities, and provide a metric to compare these abilities across models and versions.**

***Index Terms*—LLM, linguistics, syntax, ambiguity, recursion, movement, phonology, theory, rule, OpenAI, GPT, o1, Meta, Llama, transformers, generalization, metacognition.**

## 1. Introduction

We investigate the theoretical linguistic abilities of four large language models (LLMs): OpenAI's GPT-3.5 Turbo (Brown et al., 2020), GPT-4 (OpenAI, 2023), and o1 (OpenAI, 2024), as well as Meta Llama 3.1 (Dubey et al., 2024). We find that OpenAI o1 vastly outperforms other models on a variety of linguistic analysis tasks, being able to generate correct syntactic and phonological analyses within some of the popular approaches in modern linguistics (e. g. Chomsky, 1993; Chomsky and Halle, 1968). This suggests that OpenAI o1 may be the first large language model with metalinguistic abilities, i. e. with abilities not only to use language, but also to explicitly reason about it. We speculate that o1's advantage stems from its chain-of-thought mechanism which mimics the structure of reasoning human linguists use to solve analytical problems.

G. Beguš is with the Department of Linguistics, University of California, Berkeley, Berkeley, CA 94720, USA (e-mail: begus@berkeley.edu) and Project CETI, New York, NY and Roseau, Dominica.

M. Dabkowski is with the Department of Linguistics, University of California, Berkeley, Berkeley, CA 94720, USA (e-mail: dabkowski@berkeley.edu)

R. Rhodes is with the Center for Cognitive Science, Rutgers University, New Brunswick, NJ 08901, USA (e-mail: ryan.rhodes@rutgers.edu)

In the rest of the paper, we summarize previous findings on the linguistic abilities of large language models, outline a methodology for testing their metalinguistic abilities, and present several case studies in syntax and phonology, which include distinguishing between different readings in ambiguous structures, identifying different types of linguistic recursion and syntactic movement, generating tree diagrams that illustrate the analyses, and solving phonological datasets. Finally, we discuss future directions for this line of inquiry.

## 2. Background

A large body of work has tested the linguistic abilities of neural networks trained on text, ranging from recurrent neural networks (RNNs) (Gulordava et al., 2018; Matusevych et al., 2022), to long short-term memory networks (LSTMs) (Linzen et al., 2016), to—most recently—*transformers* (deep neural networks for sequential data processing; Vaswani et al., 2017) (Wilcox et al., 2018; Yedetore et al., 2023). The mostly transformer-based large language models have seen enormous success in recent years in a variety of fields, and ignited a debate on what their outputs mean for linguistic theory in general (e. g. Katzir, 2023; Kodner et al., 2023; Piantadosi, 2024).

However, most of the previous studies only consider the models' performance on behavioral tasks—this is to say, they test the models' ability to use language correctly. They do not examine the models' metalinguistic performance—which is the ability to generate analyses of language data. While LLMs have been shown to perform tasks such as part-of-speech (POS) tagging (T. Blevins et al., 2023), POS is only one ability and the previous models' performance is not impressive. Behzad et al. (2023) have composed the first corpus of metalinguistic questions and answers (Q&A) and used it to fine-tune LLMs to answer basic metalinguistic questions. However, even with fine-tuning, the models perform below humans per human evaluators, and the complexity of the metalinguistic answers is not at the level of formal analysis. This anticipates our finding that the earlier models, such as OpenAI GPT-3.5 Turbo (Brown et al., 2020), GPT-4 (OpenAI, 2023), and Meta Llama 3.1 (Dubey et al., 2024) are unable to perform complex formal linguistic analyses consistently and accurately across tasks. However, we show that with the most recent advances in the field, testing complex metalinguistic abilities of LLMs is now a possible line of inquiry and argue that the results of such tests can provide valuable insights into their general metacognitive abilities.[1]

## 3. Research program outline

In this paper, we advocate for exploring large language models' vast potential for testing their metalinguistic competence, including their abilities to construct linguistic analyses. While our present study focuses on syntax and phonology, one can test the performance of LLMs on any theoretical linguistic skill.

Why is this line of work important? The majority of studies thus far perform *behavioral tests* of LLMs. Behavioral tests include tasks such as asking a model whether a sentence is (un)grammatical, or seeing if a model can correctly perform a syntactic operation such as agreement, movement, or embedding (Haider, 2023). In other words, behavioral tasks test *language performance*.

Here, we outline a research program where large language models are tested on higher-level *metalinguistic abilities*. The term *metalinguistic* has several interpretations (for a detailed discussion, see Bialystok et al., 1985). We use the term *metalinguistic ability* to refer to the ability to analyze language itself and to generate formal, theoretical analyses of linguistic phenomena—simply put, to refer to the work that linguists do. Metalinguistic ability is cognitively more complex than language use (Tunmer et al., 1984); it is acquired later, and linguistic competence is its precondition. Applying a linguistic formalism from the training data to the model's own language ability in constructing an analysis is a complex metacognitive task.

Linguistic formalism presents the perfect testing ground for accessing the metacognitive abilities of large language models. We argue that this new research frontier can give us deeper insight into the LLMs' general capabilities and provide a useful metric for cross-model comparison. This line of inquiry can be understood as *behavioral interpretability* of deep learning, where the model's performance is evaluated through explicit metacognitive prompts rather than internal representations.

Many previous studies have attempted to test whether linguistic structures are learnable from surface statistics (i. e. the relative poverty of a human learner's input notwithstanding, given a sufficient quantity of input data, can the target grammar be acquired from statistical regularities alone?).[2] Large language models acquire linguistic competence from the surface statistics of their training data. Our goal is to understand whether this is a sufficient basis to analyze language itself.

Prompting a network to analyze a sentence in theoretical terms might shed light on whether the model has, at some level, access to linguistic structure that could enable or inform behavioral outputs. In other words, if the model performs well on behavioral tasks, are its outputs correct because of distributional knowledge, or are they informed by its understanding of constituency, hierarchical linguistic structure, etc.? Recent research suggests that transformers represent language hierarchically in structures that resemble syntactic trees (Murty et al., 2022), but so far such claims need to be evaluated implicitly by looking into the transformers' internal representations. The nature of the language models' internal representations presents one of the most consequential questions in the current research on generative AI. We show that LLMs can now be prompted to generate structure *explicitly* and that this may provide a useful direction for answering the overarching question.

Human linguists have arrived at a range of analytical tools that constitute the science of linguistics by reasoning about language structure from their own knowledge of their native languages (i. e., from their mental grammar). It remains to be seen whether large language models can ever achieve a similar capacity to reason from their "knowledge" about language to analyze their own grammar. In this paper, we present several case studies that suggest that the GPT architecture with chain-of-thought reasoning instantiated by OpenAI's (2024) o1 is moving in that direction.

Importantly, there are several issues that need to be kept in mind while undertaking this type of research. The first problem to consider is that of memorization. The models' training data set almost certainly included solutions to simple problem sets. Additionally, it must have included texts that contain syntactic annotations or examples (such as educational materials or linguistic research papers) for the models to be able to learn the representations used in formal linguistics. If we are interested in seeing whether an LLM has simply memorized a linguistic analysis (or a particular problem or data set), or whether it can generalize by applying the learned metalinguistic analyses to novel unobserved data, it is crucial to write new problem sets or invent wholly new languages and grammars. Doing so might also reveal whether the LLMs have acquired human-like cognitive *biases* alongside their distributional knowledge of human language.

The models tested in this paper include OpenAI's GPT-3.5 Turbo (Brown et al., 2020), GPT-4 (OpenAI, 2023), and o1 (OpenAI, 2024), as well as Meta Llama 3.1 (Dubey et al., 2024). The first two models have been chosen because of their overall popularity, the discussions they had stirred in the linguistic community (e. g. Haider, 2023; Katzir, 2023; Piantadosi, 2024), and the early reports of their successes in linguistic analysis (Beguš, 2023). OpenAI's (2024) o1 has been chosen due to its unique *chain-of-thought* architecture, and the advantage it has demonstrated in many tasks over previous models. Meta Llama 3.1 (Dubey et al., 2024) has been chosen as an open model, to serve as a control for OpenAI's close architectures whose learning might not have been fully unsupervised.

The choice of propriety models, however, raises another issue, which is that of reproducibility. Unfortunately, OpenAI's models are black boxes, and there is much we do not know about their architecture, training data, memory, or how they were optimized for human conversation. Given the opacity of the models, it will be difficult to make inferences about their abilities and internal representational states. As such, data transparency will be extremely important for reproducibility and cross-study comparison. Prompts and responses need to be disclosed in full. LLMs' outputs can be noisy and unpredictable—using the same prompt twice can generate substantially different responses. To mitigate this problem, the models' performance must be measured on several similar prompts or several times on the same prompt.

[1] We do not aim to argue for or against specific claims in linguistics. Rather, we simply recognize that theoretical linguistics constitutes one approach to analyzing natural language, and since LLMs are trained primarily on language data, linguistic theory allows exploring the extent of LLMs' higher-order metacognitive abilities.

[2] We recognize that the learning environments for large language models and human learners are very different. LLM input is both more impoverished—it consists only of text and lacks e. g. the sensory information embedded in phonology, prosody, gesture, etc.—and richer, in the sense that LLMs are trained on orders of magnitude more examples than human learners will ever experience. However, there is still a question of whether natural grammars are in principle learnable from statistics alone.

Together with the paper, we release our dataset that can be used to evaluate other models on the same tasks. The dataset (Beguš, Dabkowski, et al., 2025) includes 30 new sentences for each of the three syntactic tasks (90 sentences in total) and 30 phonological problems, each with 40 invented words showing a different phonological process. We anticipate our data set to be useful until new model versions are trained on it. More broadly, Beguš, Dabkowski, et al. (2025) can serve as a model for constructing novel datasets to test LLMs' metalinguistic abilities.

Finally, we note that the metalinguistic abilities of LLMs can be useful in the classroom, especially in linguistics education. Many instructors have already begun incorporating GPT models into their coursework, typically by using them to generate outputs that students can then critique or improve. While o1's performance far exceeds that of other models on many tasks, it is still often far from perfect, allowing students to take the role of the instructor and critically assess the model's mistakes. This can teach the students how to evaluate and interpret models' performance—a task that we believe will be increasingly important in the future.

## 4. Methods

In this section, we describe the methods we have used and motivate our experimental design choices.

We ran four experiments, prompting LLMs to analyze ambiguous structures (§5-A), linguistic recursion (§5-B), syntactic movement (§5-C), and phonological generalization (§5-D).

For each experiment, a single templatic prompt was designed. Depending on the experiment, the prompt asked the models to do a single task (§5-C) or several subtasks. The templatic prompt was then completed with 30 different test sentences (or test data sets for the phonological task in §5-D), yielding 30 test items per experiment ($4 \times 30 = 120$ test items in total). To ensure that the models were applying their "knowledge" of theoretical frameworks to novel problems (as opposed to regurgitating analyses present in the training data), all the test sentences (and the phonological datasets in §5-D) were novel (i. e. constructed for the purpose of this paper and not available online).

The test data set was constructed to represent a scattershot of relevant linguistic structures and patterns within each task. For example, in the movement task (subsection 5-C), two sentences involved passive voice A-bar movement, six sentences involved different types of auxiliary inversion, eight sentences focused on a variety of WH-movement constructions, etc. In the phonology task (subsection 5-D), different data sets tested the models' ability to recognize conditioning environments that precede or follow the alternating phones of interest across different natural classes of vowels, consonants, and suprasegmental features, as well as phonologically natural and unnatural, and neutralizing and non-neutralizing processes. All the categories of test items used in each task are enumerated in footnotes in the subsections of section 5. (While we have aimed at a representative coverage of different subcategories, each subcategory is represented only by a few items. As such, we do not separately analyze the models' performance on particular subcategories.)

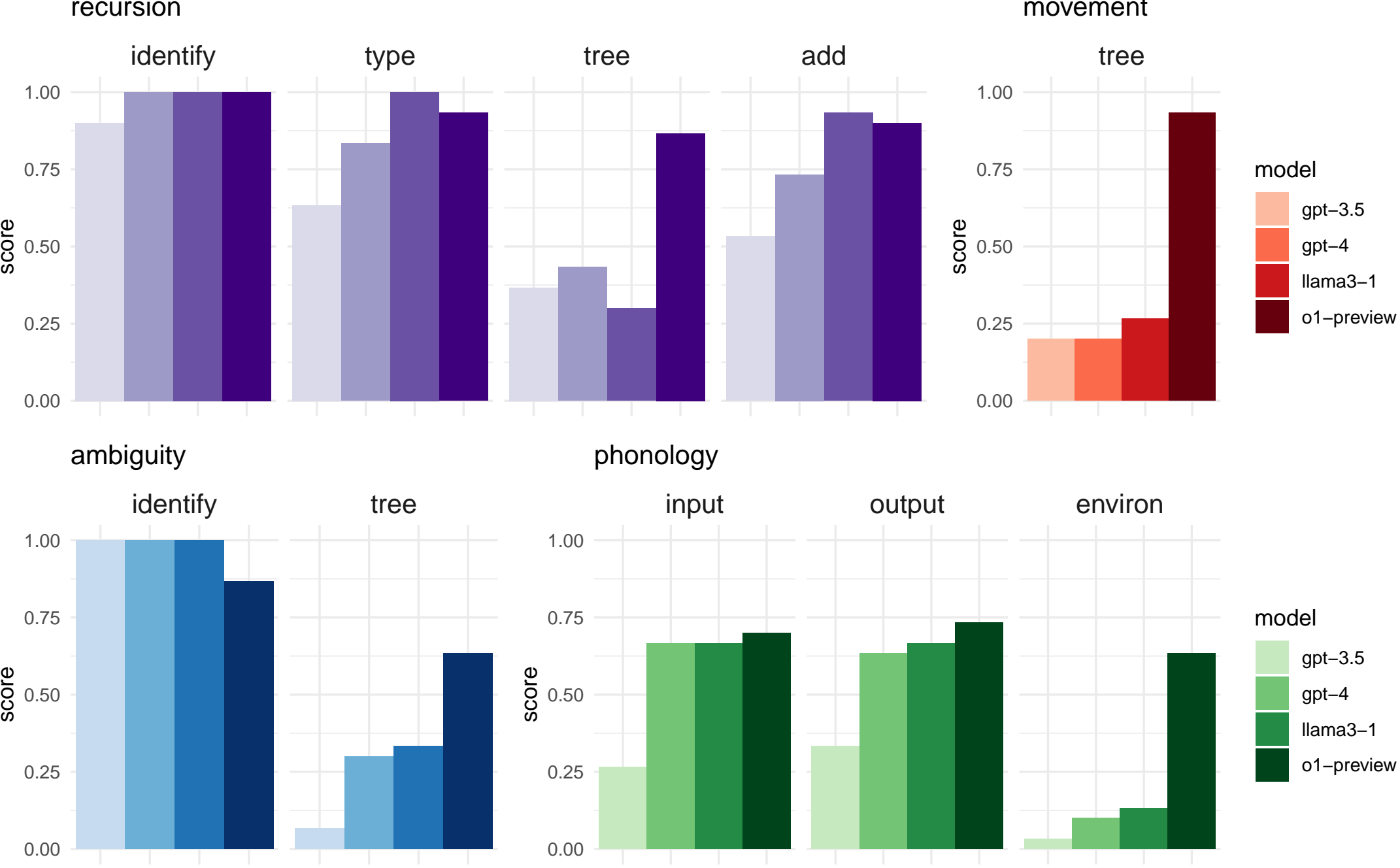

Fig. 1. Performance of the four models (`gpt-3.5-turbo-0125`, `gpt-4-0314`, `meta.llama3-1-405b-instruct-v1`, `o1-preview-2024-09-12`) on four tasks (ten subtasks in total). In the bottom left-hand quarter, the ambiguity task: (i) is the sentence ambiguous? and (ii) draw syntactic trees illustrating the ambiguity. In the upper left-hand quarter, the recursion task: (i) is the sentence recursive?, (ii) what type of recursion is it?, (iii) draw a syntactic tree representing recursion, and (iv) add a layer of recursive structure of the same type. In the upper right-hand quarter, the movement task: illustrate movement with a syntactic tree. In the bottom right-hand quarter. the phonology task: identify the phonological rule's (i) input, (ii) output, and (iii) environment.

Each test item was fed into four large language models: OpenAI's (Brown et al., 2020) GPT-3.5 Turbo (specifically `gpt-3.5-turbo-0125`), OpenAI's (2023) GPT-4 (`gpt-4-0314`), Meta's (Dubey et al., 2024) Llama 3.1 (`meta.llama3-1-405b-instruct-v1`), and OpenAI's (2024) o1 (`o1-preview-2024-09-12`). Thus, in total, we produced $120 \times 4 = 480$ model outputs.

We used default temperature settings—this is to say, 0.5 for Meta's `meta.llama3-1-405b-instruct-v1` and 1.0 for the OpenAI models (i. e. `gpt-3.5-turbo-0125`, `gpt-4-0314`, and `o1-preview-2024-09-12`). Note that within the range of 0.0–1.0, changes in temperature do not affect the LLMs' problem-solving ability (Renze et al., 2024).

Since the measure of success on a linguistic analysis task is—to a certain extent—subjective, the models' replies were independently evaluated ("graded") by three graduate linguistics students who were compensated for their work. Each student had previously taken classes in phonology and syntax, giving them the prerequisite knowledge of relevant theoretical frameworks. Each grader was instructed to evaluate the subtask in a binary fashion—as either correct (1) or incorrect (0).[3] Each subtask (10 total) includes 30 test items. Each item was tested on 4 models and evaluated by 3 students. As such, we generated $10 \times 30 \times 4 \times 3 = 3600$ data points in total.

In Figure 1 (and in the following discussion), we show the models' performance as determined by majority vote—i. e. a subtask was counted as completed correctly (1) only if two or three of the three students graded it as such. (In section 6, we conduct a statistical analysis, where each evaluator's grade is treated as a separate measurement.[4])

All of our data is publicly accessible and available for download as Beguš, Dabkowski, et al. (2025).

## 5. Results

### A. *Ambiguous structures*

The first three experiments tested the models' performance on tasks related to syntax—a subfield of linguistics that studies the structure of sentences and the rules or constraints by which sentences are derived.

In the first task, we asked the models to identify whether a sentence was syntactically ambiguous (all the test sentences were) and to illustrate the ambiguity with tree diagrams that used representations assumed by *X-bar theory* (Chomsky, Jacobs, et al., 1970) and can be rendered with LaTeX's `forest` package (Živanović, 2017). X-bar theory is a syntactic framework which represents sentences as made up of uniform, hierarchically structured phrases.[5] This allows for modeling linguistic facts such as structural ambiguity.

An example prompt testing an ambiguity between a possessed noun reading and a subject-plus-participle reading is given in (1). The other 29 test items differed from the example text prompt only in the last test sentence.[6]

(1) PROMPT

```
Consider the sentence below. Is it
syntactically ambiguous? If so, provide code
for syntactic trees representing each possible
parse of the sentence that can be rendered
with LaTeX's forest package. Assume X-bar
theory.
"Eliza wanted her cast out."
```

The replies of the four models are given below.[7] All the models identify the given sentence (*Eliza wanted her cast out*) as ambiguous. However, only o1 properly illustrates the nature of the ambiguity with syntactic trees. In one parse of the sentence, *her* is a possessor in the small clause subject *her cast*, and *out* is a predicate. In the other parse, *her* is the subject of a small clause and *cast out* is the verb phrase. While some of the model's choices (such as labeling *out* as an adjective) are non-standard and parts of the syntactic structure could be fleshed out further (e. g. the model does not represent the internal structure of the phrasal verb *cast out*), the analysis is overall correct. An abridged output of the model is given in (2).

(2) o1's REPLY

```
Yes, the sentence "Eliza wanted her cast out."
is syntactically ambiguous. The ambiguity lies
in whether "her cast" functions as a single
noun phrase (e.g., a medical cast that belongs
to her) with "out" as a predicate adjective,
or whether "her" is a pronoun object of
```

[3]The graders were instructed to focus only on the specific skill tested in each subtask, such that the models were not penalized for mistakes irrelevant to the "competency" under investigation. For example, if a model is being tested on its ability to represent ambiguity with tree diagrams, it would not be marked down for showing V-to-T movement (generally assumed to be incorrect for English). For the full project manual detailing the instructions given to the graders, see Beguš, Dabkowski, et al. (2025).

[4]In the statistical analysis, we excluded the two *identify* subtasks, which left us with $8 \times 30 \times 4 \times 3 = 2880$ data points.

[5]In maximally fleshed-out structures, each phrase consists of the core word a.k.a. *head* (X), the *intermediate projection* ($X'$), and the full phrase a.k.a. *maximal projection* (XP). Each node is, at most, binary branching. The head (X) determines the syntactic category of a phrase. The intermediate projection ($X'$) consists of a head, a tightly connected complement that appears closest to the head, and optional adjuncts. The *maximal projection* (XP) additionally includes an optional specifier that appears to the left of $X'$.

While there are many competing syntactic frameworks, in testing the models' syntactic abilities, we used X-bar theory because its formalism is relatively well-established, prominent in the literature (hence likely present in the training data set), and straightforward to both prompt and evaluate.

[6]The 30-item set of test sentences in this task contained a variety of ambiguous structures, representing 13 different categories labeled "Poss Noun vs Subject Verb," "Poss Noun vs Subject Participle," "Poss Noun vs Subject Gerund," "Adj [Noun Noun] vs [Adj Noun] Noun," "[Mod A] N vs Mod [A N]," "DtrV NP NP vs TrV [N N]," "PP modifying VP vs Noun," "GerundVP modifying VP vs Noun," "[[N N] Mod] VP vs [N [N Mod]] VP," "[Gerund Adv] V vs Gerund [Adv V]," "[CP [Mod CP]] vs [[CP Mod] CP]," "[CP [CP Mod]] vs [[CP [CP]] Mod]," and "[[CP [CP]] conj CP] vs [CP [CP conj CP]]." All the data has been deposited as Beguš, Dabkowski, et al. (2025).

[7]The `forest` trees constructed by the older models (i. e. GPT-3.5, GPT-4, and Llama 3.1) often do not compile (or compile with errors) due to minor LaTeX mistakes. To compile the trees and show them in print, minimal alterations (such as adding missing brackets) have been made throughout the paper. Since all these altered answers are incorrect anyway, these slight modifications do not affect the potential evaluations of the models' performance. Additionally, some of the models' responses have been abridged for reasons of brevity. For full unaltered outputs, see Beguš, Dabkowski, et al. (2025).

"wanted," and "cast out" is a verb phrase (e.g., meaning to expel her).
  Below are the syntactic trees for each possible parse using LaTeX's 'forest' package, assuming X-bar theory.

**First Parse** (Where "her cast" is a noun phrase and "out" is a predicate adjective):

**Second Parse** (Where "her" is the object, and "cast out" is a verb phrase):

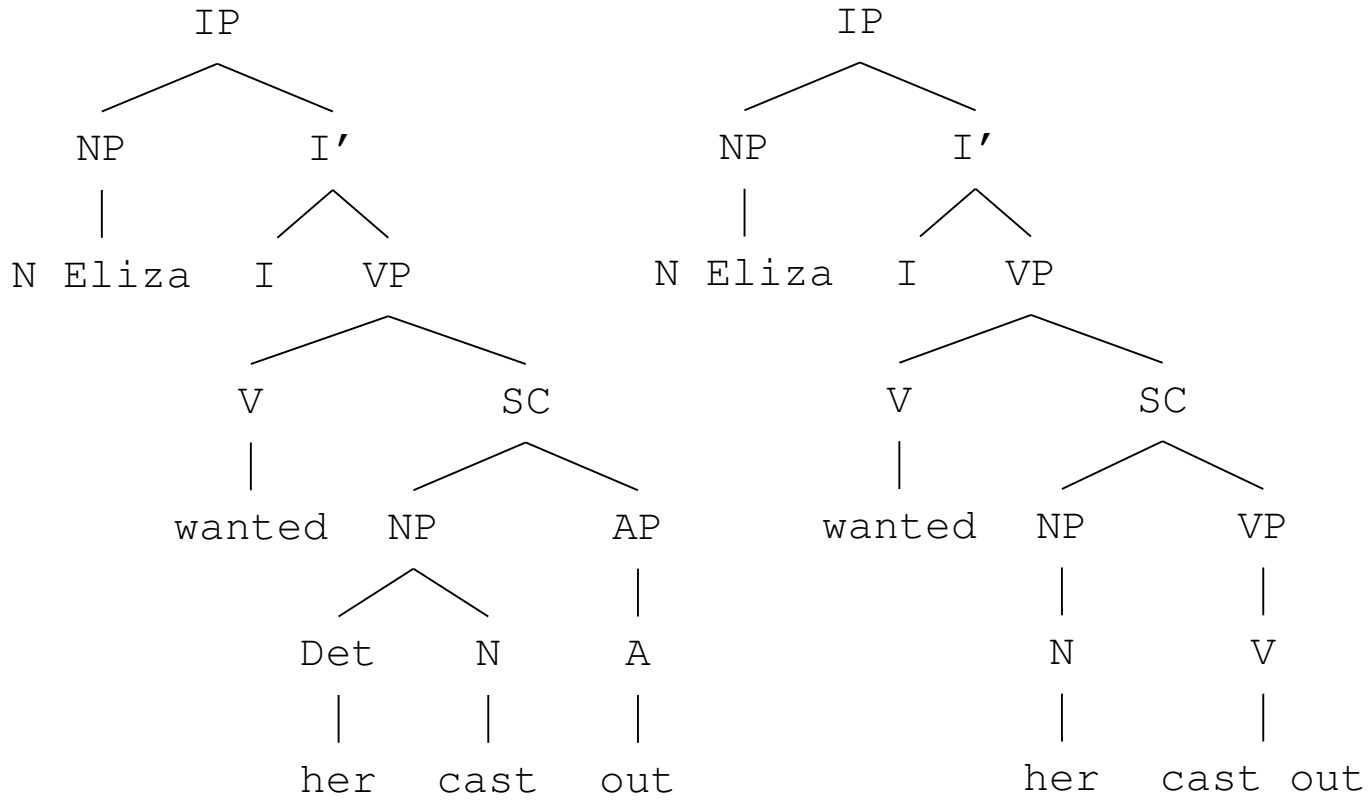

The answers provided by the other three models are unsatisfactory. The two structures outputted by GPT-3.5 Turbo roughly correspond to the two possible parses, but the grammatical categories assigned by the model are wrong—in both trees, *her cast (out)* is labeled as a noun phrase (3).

(3) GPT-3.5 Turbo's REPLY

Yes, the sentence Eliza wanted her cast out. is syntactically ambiguous.
  Here are the two possible parses of the sentence

1. Parse 1

2. Parse 2

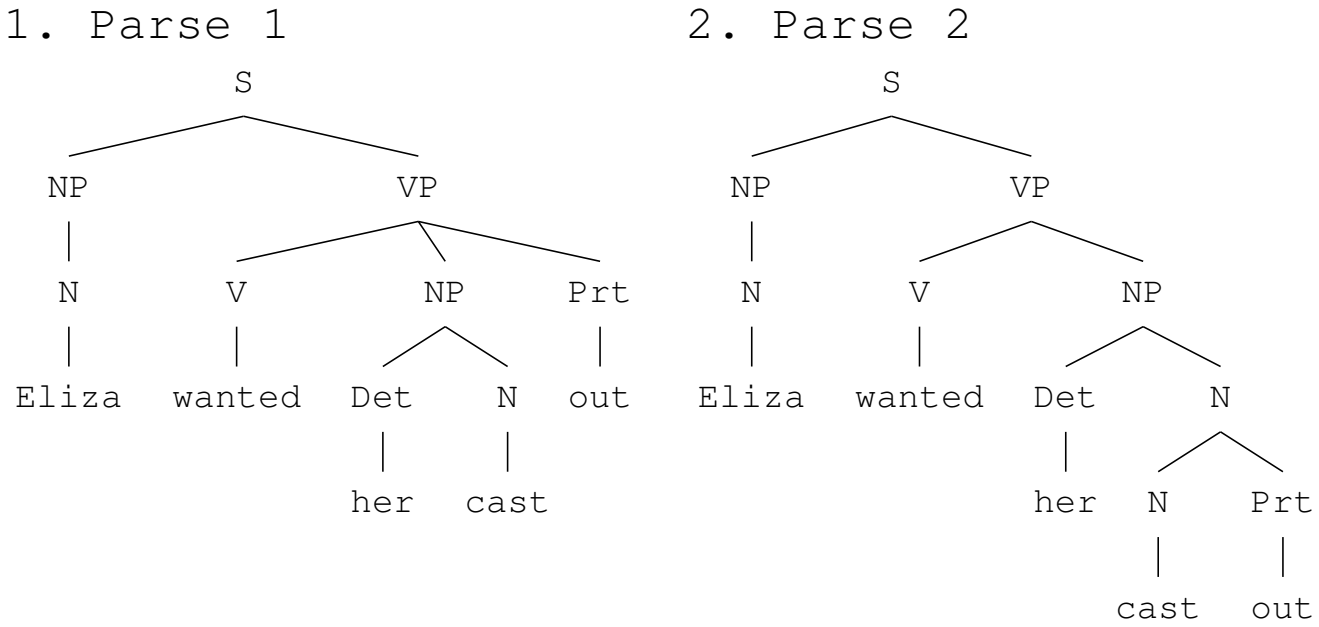

The linguistic structures given by GPT 4 are too confused to interpret. They contain non-standard syntactic labels such as NNP, B, SM, PRP, PCP, and PPartP, and the second tree is missing the small clause verb *cast* (4).

(4) GPT-4's REPLY

Yes, the sentence "Eliza wanted her cast out." is syntactically ambiguous. Here are the two possible parses for this sentence:

Parse 1: Eliza wanted [her (to be) cast out]. In this parse, Eliza wanted someone (her) to be cast out.

Parse 2: Eliza wanted (to cast) her (object) out. In this parse, Eliza wanted to cast someone (her) out.

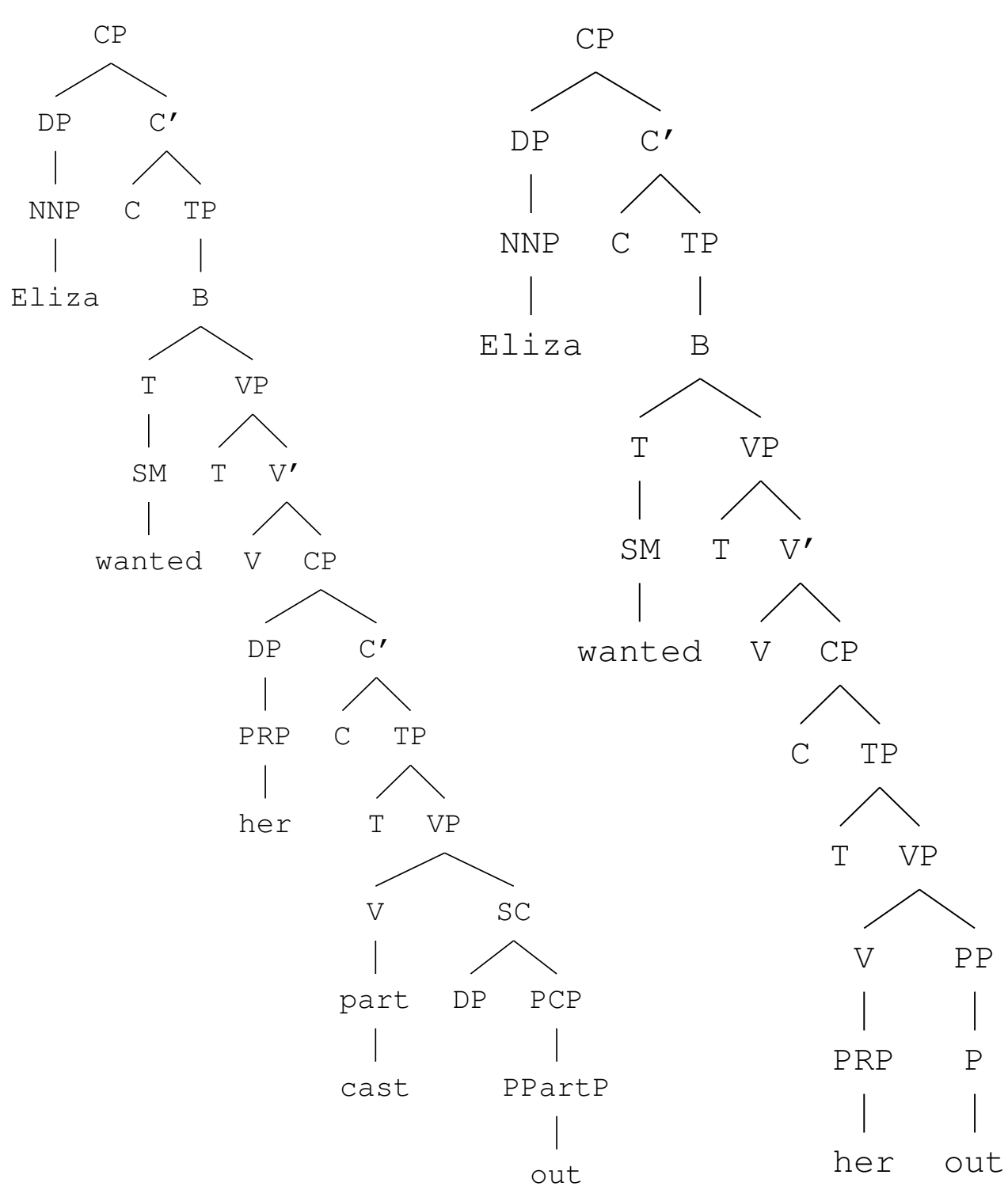

These two parses represent the syntactic ambiguity in the sentence "Eliza wanted her cast out."

Finally, Llama 3.1's tree diagrams strongly deviate from the typical assumptions about syntactic structure. For example, in both trees, *Eliza wanted* is labeled as a verb, and the second tree employs ternary branching. In addition, both trees label *cast (out)* as a verb. The model's verbal explanation of the ambiguity (provided after the tree diagrams) is also incorrect (5).

(5) Llama 3.1's REPLY

Yes, the sentence is syntactically ambiguous.
Here are the two possible parses:

**Parse 1: Eliza wanted [her cast out]**
In this parse, "her cast out" is a constituent, with "her" as the object of the verb "cast out".

**Parse 2: Eliza wanted [her] [cast out]**
In this parse, "her" is the object of the verb "wanted", and "cast out" is a separate constituent.

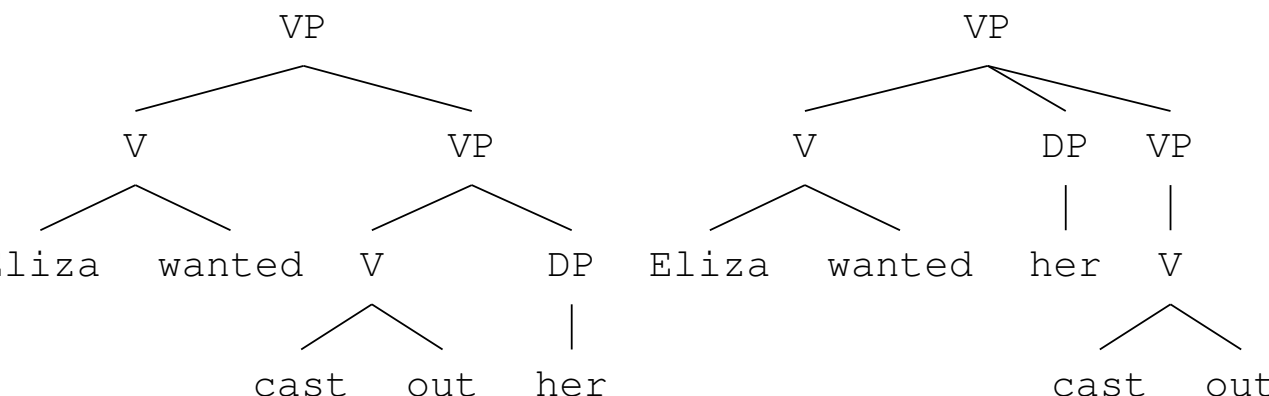

  Note that the first parse implies that Eliza wanted someone to cast her out, while the second parse implies that Eliza wanted someone (her) to be cast out. The two parses have different semantic interpretations.

The results of the ambiguity task (showing each model's performance on 30 sentences) are graphed in the bottom left-hand quarter of Figure 1. By and large, all four models correctly identify the sentences as ambiguous. However, o1 scores 0.63 on the tree generation task, considerably outperforming other models, which score at or below 0.33.

### *B. Linguistic recursion*

Next, we tested the models' ability to perform tasks related to linguistic recursion. A recursive process or pattern is one that repeats in a self-similar or self-referential manner. In linguistics, recursion refers to the embedding of phrases within phrases of the same type. This provides linguistic units with potentially infinite layers of depth and complexity. Recursion is arguably one of the few properties unique to human language (Hockett, 1960). For this reason, it has been one of the most widely studied, but also hotly debated aspects of language (Chomsky, 2014; Hauser et al., 2002 vs. Everett, 2005; Jackendoff et al., 2005).

In this task, we prompted the models to (i) say whether a given sentence is recursive, (ii) identify the recursive part and say what type of recursion it contains, (iii) draw a syntactic tree representing the sentence, and (iv) add another layer of recursion of the same type. The set of test sentences contained a variety of ambiguous structures, including adjectival recursion, possessive recursion, prepositional phrase (PP) recursion, and clausal (CP) recursion. Within the causal recursion category, some items tested the models' performance on center-embedding structures with one and two levels of embedding. For all the prompts and replies, see Beguš, Dabkowski, et al. (2025).

Two example items testing the models' performance on adjectival recursion (6) and doubly center-embedded clausal recursion (8) are given below. Due to space considerations, only o1's replies are reproduced (7, 9).

(6) PROMPT

```
Consider the sentence below and complete the
following three tasks:
1. Does the sentence in question contain
an instance of recursion? If so, identify
the recursive part and say what kind of
recursion it is. Note that there are
different types of recursion, e.g. adjectival
recursion (an adjective modifying an
adjective-modified noun), Saxon Genitive
a.k.a. possessive ’s recursion (an
’s-possessed ’s-possessor), prepositional
phrase recursion (a prepositional phrase with
another prepositional phrase inside), clausal
recursion (a clause within another clause),
and so on.
2. Provide code for a syntactic tree
representing the structure of the sentence
that can be rendered with LaTeX’s forest
package. Assume X-bar theory.
3. If you identified that the sentence
contains a recursive structure, expand it by
adding another layer of recursion (of the same
type) to the identified structure.
"Unidentified flying objects may have
conflicting characteristics."
```

Given the above prompt, o1 correctly identifies the sentence as recursive, the recursion as adjectival, and draws a tree that shows a noun recursively modified by an adjective. When asked to add another layer of recursion, the model adds *recently sighted* between *flying* and *unidentified* (7). Thus, o1 performs correctly on all subparts of this adjectival recursion task.

(7) o1's REPLY

**Task 1: Identifying Recursion in the Sentence**
Yes, the sentence contains an instance of recursion. The recursive part is the noun phrase **"Unidentified flying objects"**, which exhibits **adjectival recursion.** In this structure:
- The adjective **"unidentified"** modifies the noun phrase **"flying objects".**
- Within **"flying objects"**, the adjective **"flying"** modifies the noun **"objects".**

This creates a recursive adjectival modification where an adjective modifies a noun that is already modified by another adjective.

---

**Task 2: Syntactic Tree in LaTeX with the Forest Package (Assuming X-bar Theory)**
Below is the code for a syntactic tree representing the structure of the sentence, using LaTeX's `forest` package and assuming X-bar theory:

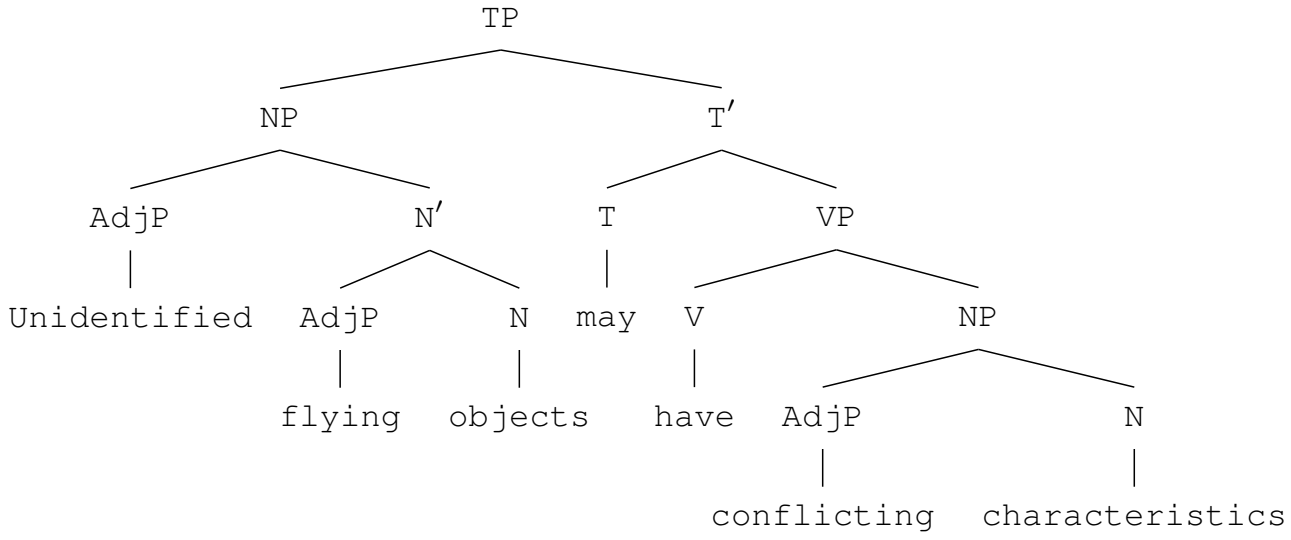

---

**Task 3: Expanding the Sentence with Another Layer of Adjectival Recursion**
Adding another layer of adjectival recursion to the identified structure, we can expand the sentence as:
**"Unidentified recently sighted flying objects may have conflicting characteristics."**
In this expanded noun phrase:
- The adjective **"unidentified"** modifies **"recently sighted flying objects".**
- The adjective **"recently sighted"** modifies **"flying objects".**
- The adjective **"flying"** modifies **"objects".**

This adds another layer of adjectival recursion, where adjectives recursively modify noun phrases that themselves contain adjectives.

An example center-embedding sentence is given in (8). Center embedding refers to a specific type of structure where a phrase is nested in the middle of another phrase of the same type. Center-embedding sentences are difficult to process, resulting in an impression of ill-formedness. Thus, although grammatical, center-embedding sentences are virtually unattested in natural discourse. This means that LLMs have almost certainly not been exposed to center-embedding structures at a large scale. As such, they are a useful test for the models' generative syntactic ability.

(8) PROMPT

```
[same instructions as in (6)]
```

```
"The worldview that the prose Nietzsche wrote
expressed was unprecedented."
```

Confronted with the above sentence, o1 correctly identifies the sentence as containing clausal recursion. The model draws a tree that captures the structure of center embedding and adds a third layer of center embedding by sandwiching *Nietzsche* between *the philosopher* and *admired* (9).

(9) o1's REPLY

**Answer to Task 1:**

Yes, the sentence contains an instance of recursion. Specifically, it involves **clausal recursion**, where a relative clause contains another relative clause. The recursive part is within the noun phrase "the worldview that the prose Nietzsche wrote expressed." Here, the noun "worldview" is modified by the relative clause "that the prose Nietzsche wrote expressed," which itself contains the embedded relative clause "Nietzsche wrote" modifying "prose." The structure can be illustrated as:

- **Main Clause:** The worldview ... was unprecedented.
  - **First Relative Clause:** that the prose **[Second Relative Clause]** expressed
    - **Second Relative Clause:** [that] Nietzsche wrote

**Answer to Task 2:**

Below is the code for a syntactic tree representing the structure of the sentence, using LaTeX's **forest** package and assuming X-bar theory. This tree illustrates the recursive embedding of relative clauses within the noun phrases.

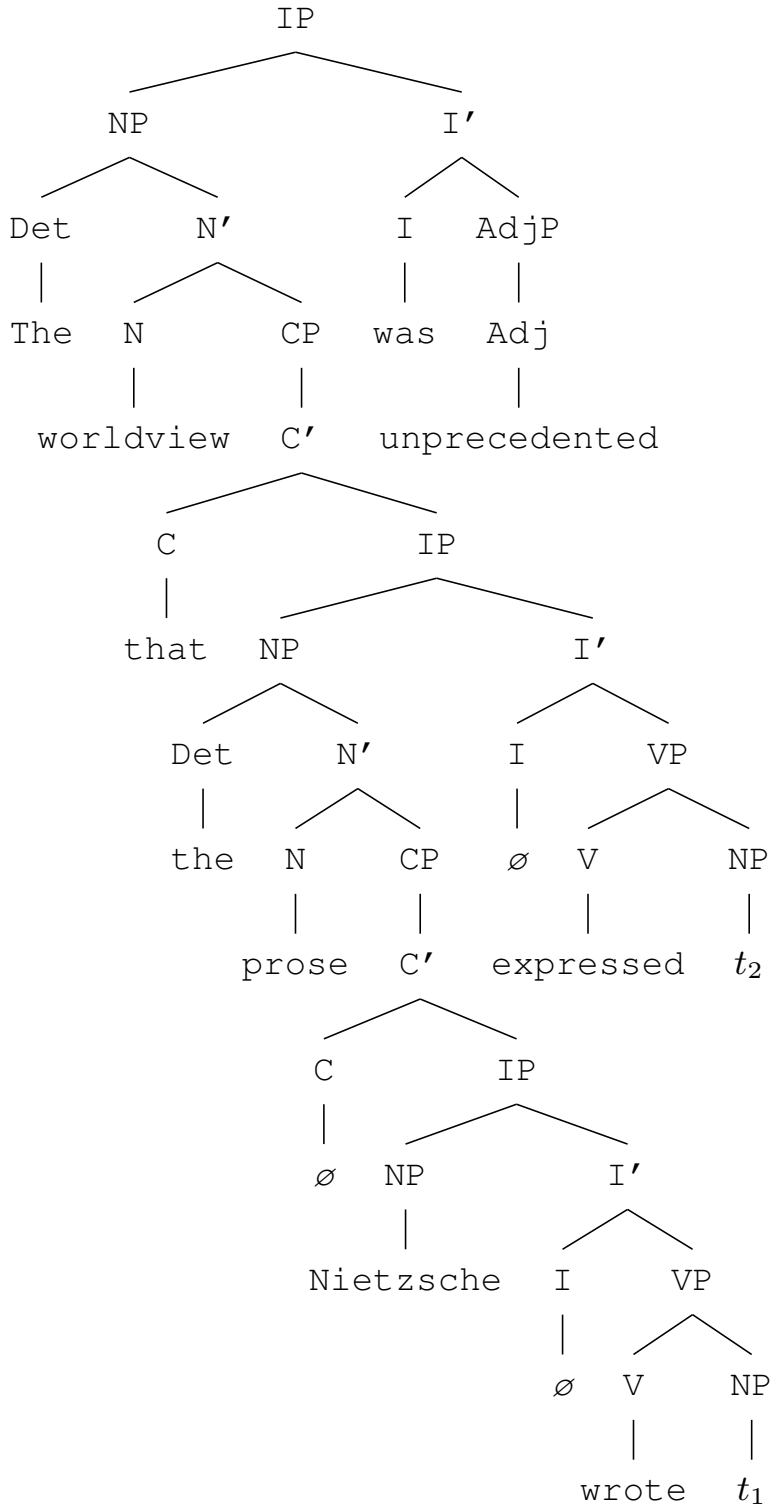

**Answer to Task 3:**

Expanding the sentence by adding another layer of clausal recursion results in:

**"The worldview that the prose that the philosopher Nietzsche admired wrote expressed was unprecedented."**

The overall results of the recursion task (showing each model's performance on 30 sentences) are graphed in the top left-hand quarter of Figure 1. By and large, all four models correctly identify the presence of recursion in the test sentences. All four models perform reasonably well on the tasks of identifying the recursion type and adding a recursive layer, with their performance ranging between 0.53 and 1.0. An incremental improvement on both tasks is seen in the family of OpenAI's models across generations (from GPT-3.5 Turbo to GPT-4 to o1). However, on the task of recursive tree generation, o1 scores 0.87, vastly outperforming the other three models, which score ∼0.36.

### C. Syntactic movement

Then, we tested the models on syntactic movement. In theoretical linguistics, movement is an operation whereby a more deeply embedded constituent is placed in a higher position in a sentence. The original position of the moved constituent is referred to as its trace.[8]

In this task, we asked the models to draw syntactic trees for the test sentences, explicitly representing movement with traces, i. e. by entering a `t` in the original position of the moved constituent.[9] An example prompt with WH-movement past a C-boundary is given in (10).[10]

(10) PROMPT

```
Consider the sentence below. Provide code for
a syntactic tree representing its structure
that can be rendered with LaTeX's forest
package. Assume X-bar theory. Pay special
attention to syntactic movement and represent
all instances of movement with (indexed)
traces. Do not write code to draw the actual
syntactic arrows that connect the traces.
"When do you think she will arrive?"
```

The replies of the four models are given below. Only o1 properly illustrates the movement of *when* from the end to the front of the sentence (11). (The model even shows the intermediate landing site of *when* in C between *think* and *she*.)[11]

(11) o1's REPLY

```
The syntactic tree for the sentence "When do
you think she will arrive?", represented in
```

[8]In minimalism (Chomsky, 1993, 1995, 2000), movement is reanalyzed as internal merge, and traces are considered to be unpronounced copies of the higher remerged constituents; the two analyses are functionally equivalent for the purpose of our task.

[9]In human-drawn tree diagrams, movement is also typically illustrated with arrows linking the trace and the landing position. However, our pilot tests revealed that the models were rarely able to add arrows to the `forest` trees without generating LaTeX syntax errors. As such, we explicitly prompted the models *not* to draw arrows.

[10]The full set of 30 sentences tested different types of movement, including passivization, auxiliary inversion in polar questions, conditionals, and negative/only sentences, WH-questions, including WH-questions with pied-piping, P-stranding, and past a C-boundary, topicalization, locative fronting, raising to subject by adjectives and by verbs, raising to object, tough-movement, and movement in relative clauses. For all the prompts and the models' outputs, see Beguš, Dabkowski, et al. (2025).

```
LaTeX's forest package using X-bar theory and
including movement traces, is as follows:
```

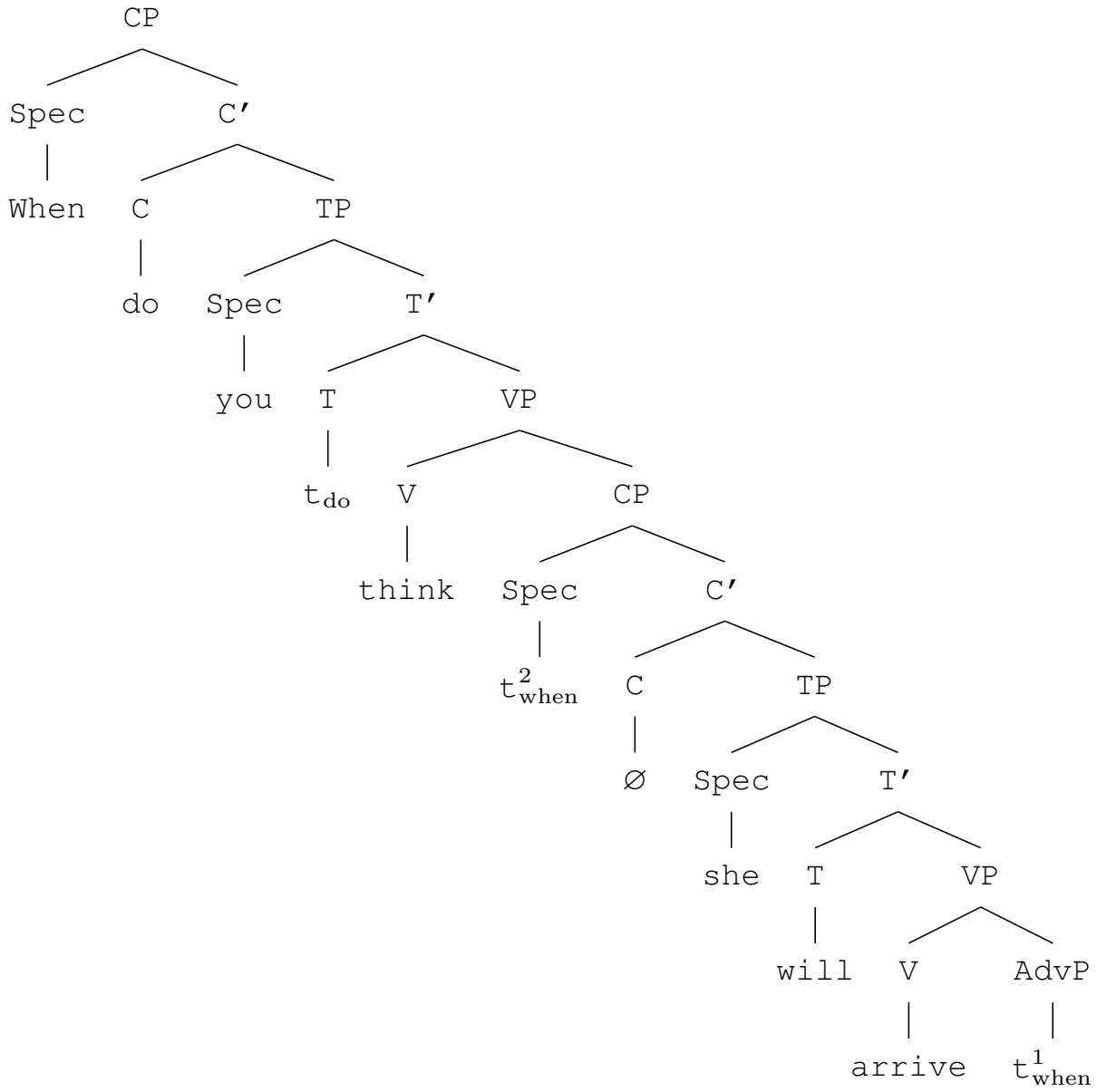

GPT-3.5 constructs a technically valid—though practically too schematic—diagram of the sentence, where any overt presentation of movement is absent (12). The model's output is evaluated as incorrect.

(12) GPT-3.5 Turbo's REPLY

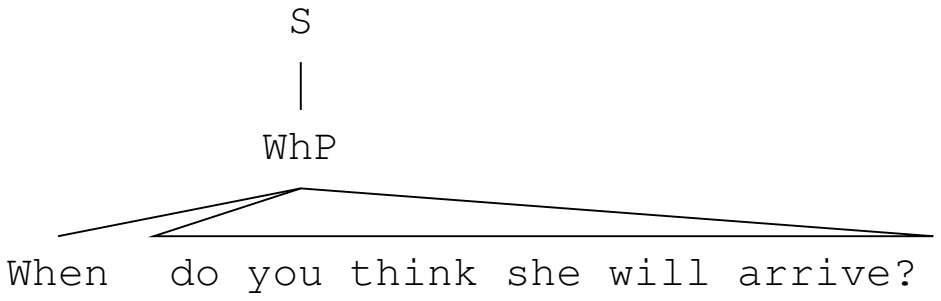

GPT-4 appends a LngP *think she* at the end of the sentence (13). Since that constituent was absent from the input, the model's output represents a failure of analysis.

(13) GPT-4's REPLY

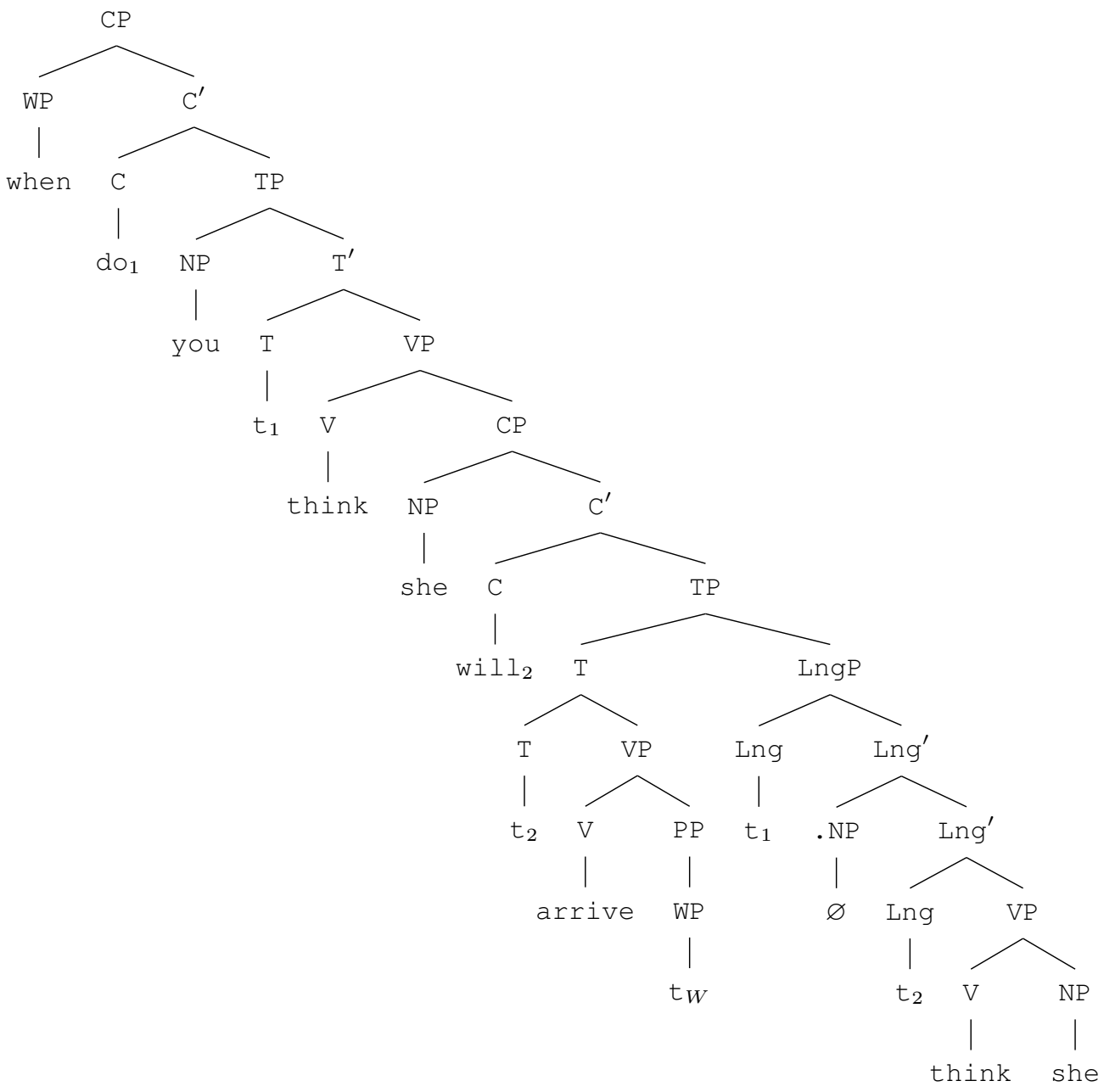

Finally, Llama 3.1 scrambles the sentence into *do will arrive she when think will arrive she you*, again resulting in failure (14).

(14) Llama 3.1's REPLY

```
Here is the code for the syntactic tree of the
sentence "When do you think she will arrive?"
using LaTeX's forest package, following X-bar
theory and representing syntactic movement
with indexed traces:
```

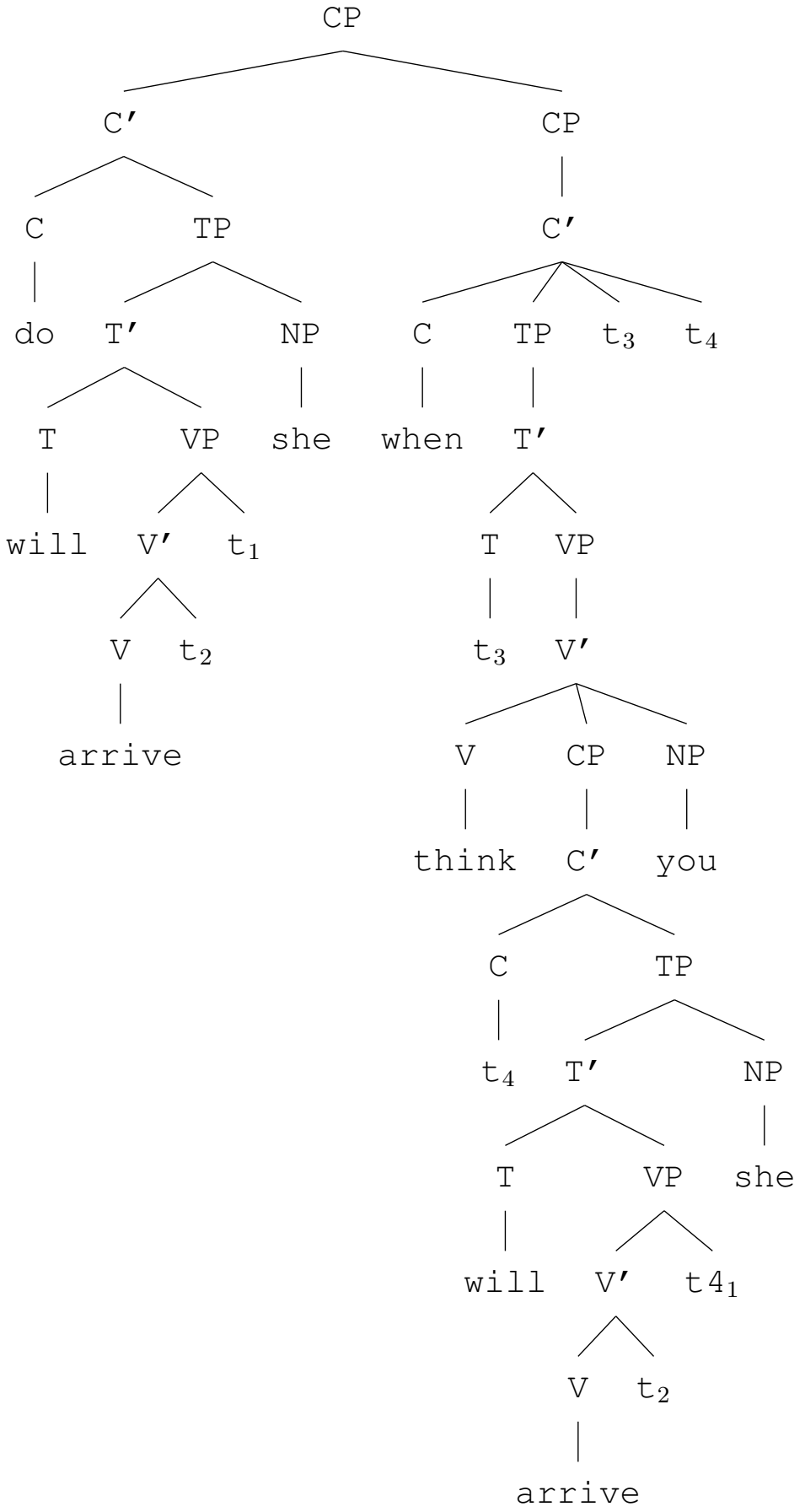

The results of the movement task are graphed in the top right-hand quarter of Figure 1. OpenAI's (2024) o1 scores 0.93, leaving the other three models (GPT-3.5 Turbo, GPT-4, and Llama 3.1) behind, which score 0.2, 0.2, and 0.27, respectively.

### D. Phonological rules

Finally, we tested the four models on a phonology task. Phonology is a subfield of linguistics that studies the organization of sounds within a language. In phonological theory, the facts of sound distribution or alternation can be modeled as rules transforming one sound into another (e. g. Chomsky and Halle, 1968; Vaux, 2008) or as constraints whose relative ranking determines the "best" form of a word (e. g. Kager, 1999; Prince et al., 1993).

[11] In addition, o1 also correctly represents the T-to-C movement of *do*. However, only one type of movement was evaluated per task. In this task, we graded the models' performance in capturing WH-movement. As such, a model's success or failure in capturing the T-to-C movement was immaterial to our assessment.

In this task, we asked the models to write a phonological rule that captures the distributional properties of specified phones in a data set. Each data set represented a toy language constructed specifically to illustrate a particular phonological process, and consisted of 40 invented "words." An example (abridged) prompt with a dataset showing intervocalic fricative gemination is given in (15).[12] The other 29 test items had different datasets (and correspondingly, different phones were identified as phones of interest).[13]

(15) PROMPT

Below, you are given a list of 40 words, which are the surface forms in a language you have not previously encountered. Each surface form is given as a string of phones separated by spaces. What phonological process operating in this language can you observe? Focus on the following phones of interest: f v s z ʃ ʒ θ ð x ɣ h ɦ ff vv ss zz ʃʃ ʒʒ θθ ðð xx ɣɣ hh ɦɦ
State the phonological process as a rule such as e.g. A → B / _ C, A → B / C _, or A → B / C _ D, where A stands for the underlying phoneme, B for its surface realization, and C (and D) specify the environment where A is realized as B. State your rule in the most general possible terms, i.e. refer to natural classes whenever possible (instead of simply listing the affected phonemes).
k u r x
a b ɣ a
a ɦ t a p o
a f f o
u v v i l i s
p o z z i j a
w i ʃ ʃ i t
v o l u ʒ ʒ i p
a m e s s i
h a o
... [30 more words]

[12]We inserted a space between each two characters in all the test words to increase the probability of each character being tokenized separately. A previous pilot study we conducted showed that spacing the characters out improves the models' performance.

The models often correctly identify the input and output of the phonological rules. (This might be due to the fact that all phones of interest were already identified in the prompt.) However, only o1 recognizes the environments of rule application in the majority of tasks. For example, given the dataset above, the model correctly identifies the intervocalic gemination of fricatives (16).

(16) o1's REPLY

The observed phonological process in the given language is **fricative gemination between vowels.** That is, any fricative consonant

[13]30 problem sets tested a variety of natural phonological rules (including word-final devoicing of obstruents, nasalization of vowels after nasal consonants, pre-glottal (h ɦ ʔ) nasalization of vowels, pre-front vowel (i, e, æ) palatalization of velars, intervocalic spirantization of voiced stops, vowels (a e i o u) lengthening before approximants (a e i o u l r̩ w j), post-front vowel (i, e, æ) palatalization of alveolars, pre-nasal vowel (ã ẽ ĩ õ ũ) nasalization of voiced stops, mid vowel laxing before consonants, post-vocalic preaspiration of voiceless stops, fricativization of oral stops before oral stops, postnasal voicing of fricatives, word-final raising of low vowels, dentalization of alveolars after front vowels (i e ɛ æ), preglottalization of voiceless stops after vowels, uvularization of velars before back (o u ɔ ɑ) vowels, (post-labial consonant (m b v p f) rounding of vowels, labialization of consonants (ʷ) after rounded vowels (o y u ø ʏ ʊ), devoicing of vowels between voiceless consonants, susurration after voiced obstruents, raising of mid vowels before voiced consonants, and pre-vocalic aspiration of voiceless stops), and some unnatural phonological rules (including word-final voicing of obstruents, denasalization of vowels after nasal consonants, high vowel lowering after laterals (l ʎ ɫ ɬ ɮ), and gemination of fricatives between vowels; for a discussion of unnatural rules, see Beguš, 2018, 2020, 2022; J. Blevins, 2004; Kiparsky, 2006). We varied whether some of the rules were neutralizing or non-neutralizing. For all the data sets and the models' solutions, see Beguš, Dabkowski, et al. (2025).

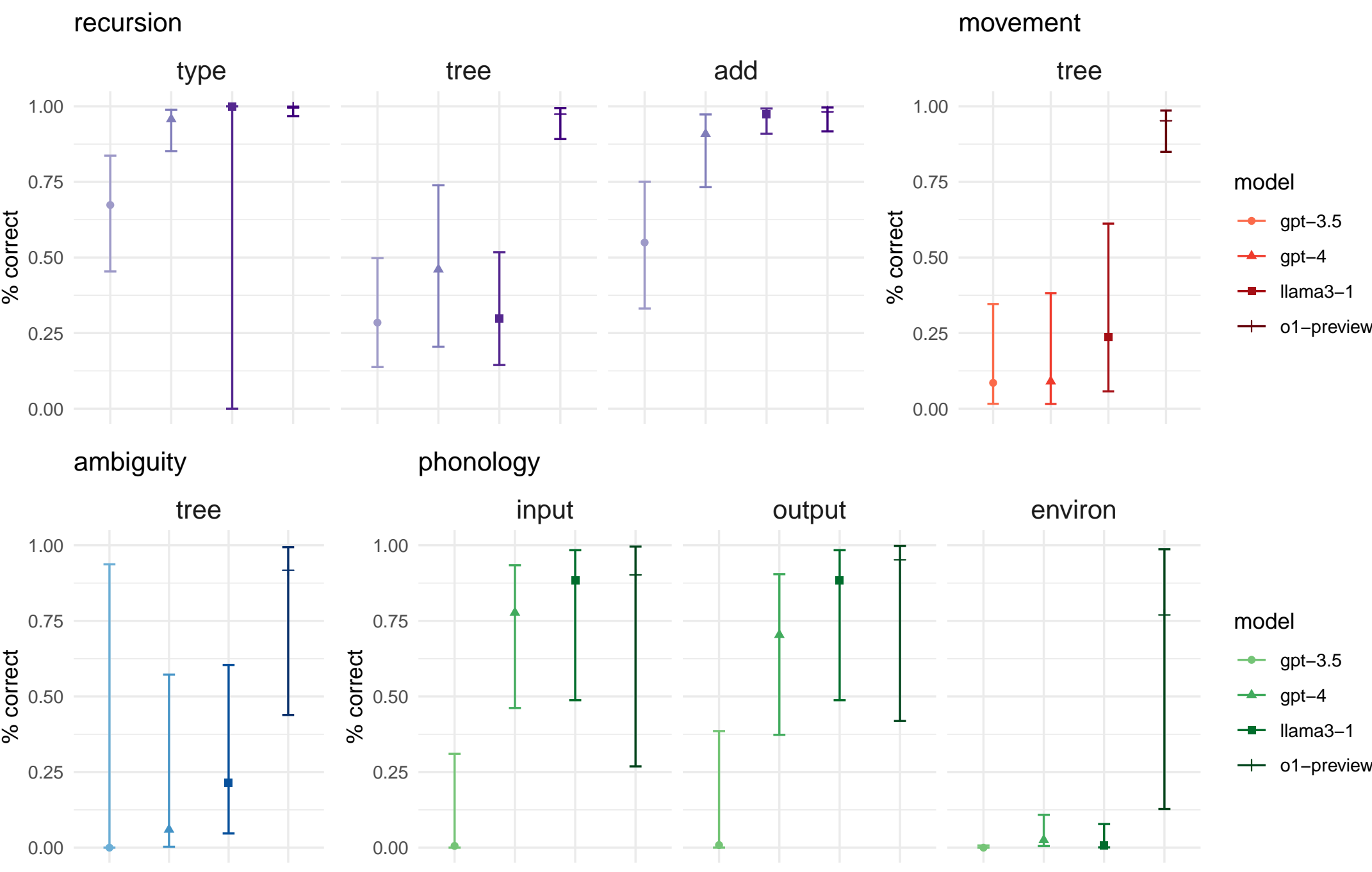

Fig. 2. Estimates with 95% CI of the subtasks in the four task groups across four evaluated models (graphed with the `effects` package; Lin, 2013).

becomes a geminate (long consonant) when it occurs between two vowels.

**Phonological Rule:**

$$[\text{+fricative}] \rightarrow [\text{+long}] \,/\, V__V$$

This means that any fricative consonant (denoted by $[+fricative]$) becomes lengthened (denoted by $[+long]$) when it is in the environment between two vowels (Vowel_Vowel).

By comparison, GPT-3.5 Turbo does not recognize gemination, does not observe that the rule applies between vowels, and does not generalize across subcases involving different fricatives (17).

(17) GPT-3.5 Turbo's REPLY

θ → f / _ #
ð → v / # _
h → ∅ / V #
ɦ → ∅ / V #
ɣ → x / _ #
s → ʃ / V_V
z → ʒ / V_V

Rule 1: Voiceless interdental fricatives change to their corresponding labiodental fricatives at the beginning of a word.
Rule 2: Voiced interdental fricatives change to their corresponding labiodental fricatives at the end of a word.
Rule 3: Voiceless glottal fricative is deleted word-finally after a vowel.
Rule 4: Voiced glottal fricative is deleted word-finally after a vowel.
Rule 5: Velar fricative changes to velar fricative word-initially.
Rule 6: /s/ changes to /ʃ/ between vowels.
Rule 7: /z/ changes to /ʒ/ between vowels.

GPT-4's minimalist answer correctly identifies the process as that of fricative gemination. However, in the model's proposed rule, the process takes place before unstressed vowels (18). Since gemination takes place between two vowels and stress is not shown in the input dataset, the model's answer is marked as incorrect.

(18) GPT-4's REPLY

Fricative → geminate / _ V[-stress]

Finally, Llama 3.1 recognizes the process as gemination, but it does not explicitly state that it is restricted to fricatives. Furthermore, the model says that it applies prevocalically (19), which is falsified by input words such as *abɣa*, *voluʒʒip*, and *hao*.

(19) Llama 3.1's REPLY

After examining the given list of surface forms, I observed that the phonological process operating in this language is a case of consonant lengthening, specifically, a process of consonant gemination.
The rule can be stated as:

C → CC / _ V

Where C represents the consonant phonemes /f, v, s, z, ʃ, ʒ, θ, ð, x, ɣ, h, ɦ/ and V represents a vowel.
In other words, when a consonant from this set appears before a vowel, it is realized as a geminate (doubled) consonant. This rule applies to all the consonants of interest, as can be seen in the provided examples.

The results of the phonology task are graphed in the bottom right-hand quarter of Figure 1. GPT-4, Llama 3.1, and o1 score ∼0.68 on the input and output subtasks. However, o1 scores 0.63 on the environment subtask, while the other three models all score below 0.14.

## 6. Statistical Analysis

To test whether the raw data in Figure 1 show any statistical trends, we fit the data to four logistic regression mixed-effects models (Figure 2). Each annotator's score is the dependent variable in all four models. For the recursion and ambiguity tasks, we remove the *identify* subtasks as the models perform on them almost perfectly, and the lack of variation causes convergence problems. For tasks with only a single subtask (movement and ambiguity), the models have a single predictor: the model identity (reverse Helmert-coded). The models estimating the performance of tasks with more than one subtask additionally include the subtask identity as a predictor (sum-coded) with the interaction between the model and the subtask. All models include a random intercept for item (individual evaluated sentence/phonology problem) and annotator, as well as per-item and per-annotator random slopes for the model identity. The per-item random slope model identity is dropped in the movement model due to convergence issues.

On the tree drawing task for the ambiguity problem, o1 performs significantly better when compared to the mean of the other three models ($\beta = 1.9$, $z = 2.7$, $p = 0.007$). Full estimates and pairwise comparisons are given in Table I[14] and Table II.

TABLE I
Estimates of the logistic regression mixed effects model for the ambiguity task fit with the lme4 package (Bates et al., 2015). The Model predictor is reverse Helmert-coded with the order of levels: GPT-3.5, GPT-4, Llama 3.1, o1-preview.

| | Estimate | Std. Error | z value | Pr(>\|z\|) |
|---|---|---|---|---|
| (Intercept) | -3.295 | 2.02 | -1.63 | 0.1035 |
| Model1 | 4.391 | 3.62 | 1.21 | 0.2257 |
| Model2 | 1.952 | 1.24 | 1.57 | 0.1154 |
| Model3 | 1.901 | 0.70 | 2.71 | 0.0068 |

TABLE II
Pairwise comparisons of the ambiguity model with Tukey adjustment obtained with the emmeans package (Lenth, 2018).

| | contrast | estimate | SE | df | z.ratio | p.value |
|---|---|---|---|---|---|---|
| 1 | (GPT-3.5) - (GPT-4) | -8.7811 | 7.2482 | Inf | -1.211 | 0.620 |
| 2 | (GPT-3.5) - Llama 3.1 | -10.2451 | 7.1637 | Inf | -1.430 | 0.480 |
| 3 | (GPT-3.5) - O1 Preview | -13.9463 | 7.2638 | Inf | -1.920 | 0.220 |
| 4 | (GPT-4) - Llama 3.1 | -1.4640 | 1.6130 | Inf | -0.908 | 0.801 |
| 5 | (GPT-4) - O1 Preview | -5.1652 | 2.0271 | Inf | -2.548 | 0.053 |
| 6 | Llama 3.1 - O1 Preview | -3.7012 | 1.5127 | Inf | -2.447 | 0.069 |

Results are given on the log odds ratio (not the response) scale.
P value adjustment: tukey method for comparing a family of 4 estimates

[14]The LaTeX tables were generated in R 4.3.0 (R Core Team, 2024) with the xtable 1.8-4 package (Dahl et al., 2019). Their typesetting was altered manually.

On the recursion task, GPT-4 performs significantly better than GPT-3.5 ($\beta = 0.9$, $z = 2.5$, $p = 0.01$) at the means of all tasks. The estimates in Table IV show that o1 performs with the highest success on the tree drawing subtask in recursion. The pairwise comparison with Tukey adjustment suggests that o1 significantly outperforms GPT-3.5, GPT-4, and Llama 3.1. Full estimates and pairwise comparisons are given in Table III and Table IV.

TABLE III
ESTIMATES OF THE LOGISTIC REGRESSION MIXED EFFECTS MODEL FOR THE RECURSION TASK FIT WITH THE LME4 PACKAGE (BATES ET AL., 2015). THE MODEL PREDICTOR IS REVERSE HELMERT-CODED WITH THE ORDER OF LEVELS: GPT-3.5, GPT-4, LLAMA 3.1, O1-PREVIEW. THE TYPE PREDICTOR IS SUM-CODED.

| | Estimate | Std. Error | z value | Pr(>\|z\|) |
|---|---|---|---|---|
| (Intercept) | 3.496 | 158.11 | 0.02 | 0.9824 |
| Model1 | 0.874 | 0.34 | 2.54 | 0.0110 |
| Model2 | 2.365 | 210.81 | 0.01 | 0.9910 |
| Model3 | 0.256 | 52.70 | 0.00 | 0.9961 |
| type1 | -0.975 | 158.11 | -0.01 | 0.9951 |
| type2 | -3.072 | 158.11 | -0.02 | 0.9845 |
| Model1:type1 | 0.175 | 0.18 | 0.97 | 0.3315 |
| Model2:type1 | -1.576 | 210.81 | -0.01 | 0.9940 |
| Model3:type1 | 0.229 | 52.70 | 0.00 | 0.9965 |
| Model1:type2 | -0.492 | 0.19 | -2.60 | 0.0094 |
| Model2:type2 | -2.470 | 210.81 | -0.01 | 0.9907 |
| Model3:type2 | 0.813 | 52.70 | 0.02 | 0.9877 |

TABLE IV
PAIRWISE COMPARISONS OF THE RECURSION MODEL WITH TUKEY ADJUSTMENT OBTAINED WITH THE EMMEANS PACKAGE (LENTH, 2018).

| contrast | estimate | SE | df | z.ratio | p.value |
|---|---|---|---|---|---|
| type = add | | | | | |
| (GPT-3.5) - (GPT-4) | -2.0976 | 0.7845 | Inf | -2.674 | 0.0376 |
| (GPT-3.5) - Llama 3.1 | -3.4158 | 0.7991 | Inf | -4.275 | 0.0001 |
| (GPT-3.5) - o1-preview | -3.7753 | 0.8474 | Inf | -4.455 | <.0001 |
| (GPT-4) - Llama 3.1 | -1.3182 | 0.8880 | Inf | -1.484 | 0.4469 |
| (GPT-4) - o1-preview | -1.6777 | 1.0939 | Inf | -1.534 | 0.4172 |
| Llama 3.1 - o1-preview | -0.3595 | 0.9885 | Inf | -0.364 | 0.9836 |
| type = tree | | | | | |
| (GPT-3.5) - (GPT-4) | -0.7636 | 0.7495 | Inf | -1.019 | 0.7385 |
| (GPT-3.5) - Llama 3.1 | -0.0667 | 0.6470 | Inf | -0.103 | 0.9996 |
| (GPT-3.5) - o1-preview | -4.5514 | 0.8302 | Inf | -5.482 | <.0001 |
| (GPT-4) - Llama 3.1 | 0.6968 | 0.7127 | Inf | 0.978 | 0.7622 |
| (GPT-4) - o1-preview | -3.7878 | 1.0511 | Inf | -3.604 | 0.0018 |
| Llama 3.1 - o1-preview | -4.4846 | 0.8574 | Inf | -5.231 | <.0001 |
| type = type | | | | | |
| (GPT-3.5) - (GPT-4) | -2.3844 | 0.8175 | Inf | -2.917 | 0.0186 |
| (GPT-3.5) - Llama 3.1 | -20.4272 | 1897.3098 | Inf | -0.011 | 1.0000 |
| (GPT-3.5) - o1-preview | -4.4578 | 0.9607 | Inf | -4.640 | <.0001 |
| (GPT-4) - Llama 3.1 | -18.0429 | 1897.3098 | Inf | -0.010 | 1.0000 |
| (GPT-4) - o1-preview | -2.0735 | 1.2020 | Inf | -1.725 | 0.3105 |
| Llama 3.1 - o1-preview | 15.9694 | 1897.3099 | Inf | 0.008 | 1.0000 |

Results are given on the log odds ratio (not the response) scale.
P value adjustment: tukey method for comparing a family of 4 estimates

On the movement task, Llama 3.1 significantly outperforms GPT-3.5 and GPT-4 when compared to the mean of GPT-3.5 and GPT-4 together ($\beta = 0.4$, $z = 2.85$, $p = 0.004$). o1 significantly outperforms the other three models ($\beta = 1.2$, $z = 6.8$, $p = 0.0000$). All estimates and pairwise comparisons are given in Table V and Table VI.

TABLE V
ESTIMATES OF THE LOGISTIC REGRESSION MIXED EFFECTS MODEL FOR THE MOVEMENT TASK FIT WITH THE LME4 PACKAGE (BATES ET AL., 2015). THE MODEL PREDICTOR IS REVERSE HELMERT-CODED WITH THE ORDER OF LEVELS: GPT-3.5, GPT-4, LLAMA 3.1, O1-PREVIEW.

| | Estimate | Std. Error | z value | Pr(>\|z\|) |
|---|---|---|---|---|
| (Intercept) | -0.716 | 0.73 | -0.99 | 0.3239 |
| Model1 | 0.030 | 0.27 | 0.11 | 0.9100 |
| Model2 | 0.390 | 0.14 | 2.85 | 0.0044 |
| Model3 | 1.235 | 0.18 | 6.79 | 0.0000 |

TABLE VI
PAIRWISE COMPARISONS OF THE MOVEMENT MODEL WITH TUKEY ADJUSTMENT OBTAINED WITH THE EMMEANS PACKAGE (LENTH, 2018).

| | contrast | estimate | SE | df | z.ratio | p.value |
|---|---|---|---|---|---|---|
| 1 | (GPT-3.5) - (GPT-4) | -0.0604 | 0.5349 | Inf | -0.113 | 0.999 |
| 2 | (GPT-3.5) - Llama 3.1 | -1.1988 | 0.4873 | Inf | -2.460 | 0.066 |
| 3 | (GPT-3.5) - O1 Preview | -5.3593 | 0.7997 | Inf | -6.701 | 0.000 |
| 4 | (GPT-4) - Llama 3.1 | -1.1384 | 0.4928 | Inf | -2.310 | 0.096 |
| 5 | (GPT-4) - O1 Preview | -5.2989 | 0.8285 | Inf | -6.395 | 0.000 |
| 6 | Llama 3.1 - O1 Preview | -4.1605 | 0.7193 | Inf | -5.784 | 0.000 |

Results are given on the log odds ratio (not the response) scale.
P value adjustment: tukey method for comparing a family of 4 estimates

On the phonological subtasks, GPT-4 significantly outperforms GPT-3.5, Llama 3.1 significantly outperforms GPT-3.5 and GPT-4 combined, and o1 outperforms other models combined at means of all subtasks. All estimates and pairwise comparisons are given in Table VII and Table VIII.

TABLE VII
ESTIMATES OF THE LOGISTIC REGRESSION MIXED EFFECTS MODEL FOR THE PHONOLOGY TASK FIT WITH THE LME4 PACKAGE (BATES ET AL., 2015). THE MODEL PREDICTOR IS REVERSE HELMERT-CODED WITH THE ORDER OF LEVELS: GPT-3.5, GPT-4, LLAMA 3.1, O1-PREVIEW. THE TYPE PREDICTOR IS SUM-CODED.

| | Estimate | Std. Error | z value | Pr(>\|z\|) |
|---|---|---|---|---|
| (Intercept) | -1.336 | 1.04 | -1.29 | 0.1975 |
| Model1 | 3.094 | 1.12 | 2.76 | 0.0058 |
| Model2 | 1.115 | 0.38 | 2.96 | 0.0030 |
| Model3 | 1.158 | 0.34 | 3.37 | 0.0008 |
| type1 | -3.048 | 0.31 | -9.73 | 0.0000 |
| type2 | 1.427 | 0.22 | 6.52 | 0.0000 |
| Model1:type1 | 0.196 | 0.46 | 0.43 | 0.6671 |
| Model2:type1 | -0.412 | 0.29 | -1.42 | 0.1567 |
| Model3:type1 | 0.705 | 0.14 | 4.93 | 0.0000 |
| Model1:type2 | 0.096 | 0.30 | 0.32 | 0.7497 |
| Model2:type2 | 0.207 | 0.19 | 1.08 | 0.2790 |
| Model3:type2 | -0.448 | 0.12 | -3.79 | 0.0001 |

## 7. DISCUSSION

Large language models perform remarkably well on predicting the most likely tokens to generate coherent text. In this paper, we have outlined a new research program for testing the metalinguistic abilities of LLMs. We have presented several prompt designs that elicit the LLMs' performance on syntactic and phonological tasks, and believe that, if pursued, the proposed program may yield further insights into the on-going advances in generative artificial intelligence (especially in light of recent claims that emergent abilities in the large language models are a mirage; see Schaeffer et al., 2023).

TABLE VIII
PAIRWISE COMPARISONS OF THE PHONOLOGY MODEL WITH TUKEY ADJUSTMENT OBTAINED WITH THE EMMEANS PACKAGE (LENTH, 2018).

| contrast | estimate | SE | df | z.ratio | p.value |
|---|---|---|---|---|---|
| type = environ | | | | | |
| (GPT-3.5) - (GPT-4) | -6.5796 | 2.6976 | Inf | -2.439 | 0.0699 |
| (GPT-3.5) - Llama 3.1 | -5.3973 | 2.5935 | Inf | -2.081 | 0.1593 |
| (GPT-3.5) - O1 Preview | -11.4460 | 2.6874 | Inf | -4.259 | 0.0001 |
| (GPT-4) - Llama 3.1 | 1.1823 | 1.3259 | Inf | 0.892 | 0.8092 |
| (GPT-4) - O1 Preview | -4.8664 | 1.6760 | Inf | -2.904 | 0.0193 |
| Llama 3.1 - O1 Preview | -6.0487 | 1.3850 | Inf | -4.367 | 0.0001 |
| type = input | | | | | |
| (GPT-3.5) - (GPT-4) | -6.3805 | 2.1817 | Inf | -2.925 | 0.0181 |
| (GPT-3.5) - Llama 3.1 | -7.1560 | 1.9889 | Inf | -3.598 | 0.0018 |
| (GPT-3.5) - O1 Preview | -7.3518 | 2.2284 | Inf | -3.299 | 0.0054 |
| (GPT-4) - Llama 3.1 | -0.7755 | 1.1335 | Inf | -0.684 | 0.9032 |
| (GPT-4) - O1 Preview | -0.9713 | 1.6870 | Inf | -0.576 | 0.9394 |
| Llama 3.1 - O1 Preview | -0.1958 | 1.2989 | Inf | -0.151 | 0.9988 |
| type = output | | | | | |
| (GPT-3.5) - (GPT-4) | -5.6027 | 2.1474 | Inf | -2.609 | 0.0449 |
| (GPT-3.5) - Llama 3.1 | -6.7630 | 1.9544 | Inf | -3.460 | 0.0030 |
| (GPT-3.5) - O1 Preview | -7.7249 | 2.2394 | Inf | -3.450 | 0.0032 |
| (GPT-4) - Llama 3.1 | -1.1602 | 1.1275 | Inf | -1.029 | 0.7324 |
| (GPT-4) - O1 Preview | -2.1221 | 1.7294 | Inf | -1.227 | 0.6096 |
| Llama 3.1 - O1 Preview | -0.9619 | 1.3584 | Inf | -0.708 | 0.8939 |

Results are given on the log odds ratio (not the response) scale.
P value adjustment: tukey method for comparing a family of 4 estimates

More concretely, we have compared the theoretical linguistic abilities of four LLMs: GPT-3.5 Turbo, GPT-4, Llama 3.1, and o1. We have found that on many tasks, the performance of some of the models is comparable. Across all the tasks, Llama 3.1 and GPT-4 perform similarly. This result is important because while Llama 3.1 is an open-source model, OpenAI has not released details about GPT-4's training data. As such, some have speculated that GPT-4's learning might not be wholly unsupervised. Since GPT-4 does not have an obvious advantage over the open-source Llama 3.1, these speculations do not seem substantiated.

On the more difficult tasks, o1 vastly outperforms other models. Specifically, on the tree drawing tasks, o1 scores between 0.63 and 0.93, while GPT-4 and Llama 3.1 score $\sim$0.31. On the rule environment task, o1 scores 0.63, while GPT-4 and Llama 3.1 score below 0.14 (Figure 1). We speculate that OpenAI o1's unique advantage may result from the model's *chain-of-thought* mechanism, which mimics the structure of human reasoning used in complex cognitive tasks, such as linguistic analysis.[15]

Chain-of-thought reasoning involves breaking down a problem into smaller parts and using previously produced output as input for further text generation. Thus, our findings are at odds with Şahin et al. (2020), who claim that language models "lack the skill of iterative reasoning upon knowledge" (p. 1241), and challenge Katzir (2023), who shows that reprompting does not result in an improvement and argues that "further time and resources are of no use to ChatGPT" (p. 5). Instead, our results suggest that—given sufficiently enriched architecture—LLMs are capable of iterative reasoning, which is a prerequisite for complex problem solving. Our results also suggest that metalinguistic ability, one of the few uniquely human properties of language, can emerge in artificial neural models.

[15] In the first recursion subtask, OpenAI's o1 incorrectly identifies 3 out of 30 test sentences as non-recursive, doing worse than the other three models (whose performance is perfect). This is in line with previous observations that chain-of-thought reasoning may result in decreases in performance on some tasks (OpenAI, 2024).

We draw particular attention to o1's success on some of the trickier tasks. As illustrated in subsection 5-B, the model has often succeeded on complex center-embedding structures. This is notable because center embedding, while grammatical, is difficult to process. As such, center-embedding sentences are exceedingly rare and were most likely only sparsely present in the model's training data. OpenAI o1's ability to construct center-embedded sentences without being explicitly prompted to do so thus suggests that the model acquired grammatical structure beyond the simple distributional tendencies of its training data set.

OpenAI's (2024) o1 also vastly outperformed the other three models on the phonological task, providing the correct rule application environments in 19 out of the 30 cases. The model's success on this task is particularly impressive because all the datasets were constructed without reference to any natural human languages. As such, there is no possibility that the model's performance could be largely attributed to memorization. Furthermore, o1 succeeded on some rare and unnatural phonological rules, such as word-final obstruent voicing and intervocalic fricative gemination (as illustrated in subsection 5-D). Since a discussion of these kinds of processes is additionally rare or absent from linguistic literature, the model's success further hints at a general and abstract analytical ability.

Our phonological task echoes some recent work (e. g. Matusevych et al., 2022; Yedetore et al., 2023) that tests the linguistic abilities of LLMs with artificial grammar learning (AGL) paradigms (which have been used in human research for decades; e. g. Cleeremans, 1993; Perruchet et al., 2006; Reber, 1967). Nonetheless, these previous LLM studies were structured like human experiments and did not test the LLMs' metalinguistic abilities. Artificial language learning is the gold standard in LLM research because it avoids the problem of memorization. Since phonological datasets can be tailored specifically to be as *unlike* the exemplars in the training data as possible, phonology is uniquely suitable for testing the metalinguistic abilities of LLMs.

We share a public database of novel sentences that can be used for model evaluation (Beguš, Dabkowski, et al., 2025). In doing so, we provide a template for conducting further behavioral interpretability experiments on LLMs, which can be applied to other areas within linguistics, such as morphology, historical linguistics, semantics, etc.

Neural net architectures have been the subject of intense debates in cognitive science for decades. Most of the debate has centered on the capacities of artificial neural networks and their viability as models of human cognition. Proponents of symbolic processing models argue that neural networks lack systematicity, symbol grounding, meaning, abstract reasoning, and the ability to generalize beyond their training data (Bender, Gebru, et al., 2021; Bender and Koller, 2020; Berglund et al., 2024; Fodor and McLaughlin, 1990; Fodor and Pylyshyn, 1988). Proponents of connectionism have countered that neural nets can in fact replicate many core aspects of human cognition (Chalmers, 1993; Smolensky, 1988), and it has recently been argued that LLMs or other connectionist models may serve as plausible models of human cognition and brain activity (AlKhamissi et al., 2025; Beguš, 2021; Beguš and Zhou, 2022; Beguš, Zhou, and Zhao, 2023; Caucheteux et al., 2022; Goldstein et al., 2022).

The rapid advances in natural language modeling brought about by recent developments in the transformer architecture have also stirred a debate in the linguistic community. Piantadosi (2024) observes that large language models have achieved a remarkable command of language without relying on any formal linguistic representations. Consequently, he argues that LLMs invalidate Chomsky's approach to language and heralds the end of generative linguistics. Katzir (2023) rejoinders that LLMs are engineered tools, not models of human cognition, and hence do not tell us anything about how human language works. Kodner et al. (2023) add that humans acquire language with orders of magnitude less input than LLMs, so the central mystery of language acquisition remains.

Human linguists arrive at their understanding of language structure by introspection, cross-linguistic comparison, and arguments from analytical simplicity. Human judgments rely on metalinguistic cognition—a person's ability to reflect on their own language use. LLMs have been reported to give faulty grammaticality judgments, showing a *yes*-response bias and low judgment stability (Dentella et al., 2023). However, more careful probing of LLMs' internal probabilities shows that they are much better at this task than previously thought, with probabilities at or near ceiling performance on most types of sentences (Hu et al., 2024).

We introduce a new task to the discussion of LLMs' abilities—the application of explicit metalinguistics and linguistic theory. This task represents a new frontier as metacognitive abilities are generally more complex and acquired later than general cognitive functions (Flavell, 1979). Linguists have developed many analytical tools that serve to unearth the deep structure of language, such as X-bar theory and phonological rules. OpenAI o1's success at many of our tasks suggests that LLMs are moving in the direction of acquiring the same theoretical skills as human linguists. The model's limited success on center-embedding structures and phonologically unnatural datasets suggests an emergence of generalized reasoning that goes beyond the training data (also see Cho et al., 2025).

We speculate that future research might be able to leverage large language models not to supplant linguistic theory, but to help develop it in new directions. Deep neural networks have been instrumental in shrinking the hypothesis spaces and offering new solutions to problems in fields as diverse as protein design (Jumper et al., 2021), geometry (Davies et al., 2021), and cracking unknown communication systems (Beguš, Leban, et al., 2023). It remains to be seen whether any new insights can be gained by metalinguistic prompting of LLMs. With sufficiently advanced models, we might be able to test whether LLMs are capable of developing innovative theoretical solutions and proposing new analytical frameworks that have not been thought of by humans. Novel insights gained in this way would suggest that not only linguistic ability, but also *metalinguistic* ability, i. e. the ability to theorize about language itself, is not unique to humans and may arise in artificial systems such as transformers.

## References

AlKhamissi, B. et al. (2025). *The LLM Language Network: A Neuroscientific Approach for Identifying Causally Task-Relevant Units*. arXiv: 2411.02280 [cs.CL]. URL: https://arxiv.org/abs/2411.02280.

Bates, D. et al. (2015). "Fitting Linear Mixed-Effects Models Using lme4". In: *Journal of Statistical Software* 67.1, pp. 1–48. DOI: 10.18637/jss.v067.i01.

Beguš, G. (2018). "Post-nasal devoicing and the blurring process". In: *Journal of Linguistics*, pp. 1–65. DOI: 10.1017/S002222671800049X.

Beguš, G. (2020). "Estimating historical probabilities of natural and unnatural processes". In: *Phonology* 37.4, pp. 515–549. DOI: 10.1017/S0952675720000263.

Beguš, G. (2021). "CiwGAN and fiwGAN: Encoding information in acoustic data to model lexical learning with Generative Adversarial Networks". In: *Neural Networks* 139, pp. 305–325. ISSN: 0893-6080. DOI: https://doi.org/10.1016/j.neunet.2021.03.017. URL: https://www.sciencedirect.com/science/article/pii/S0893608021001052.

Beguš, G. (2022). "Distinguishing cognitive from historical influences in phonology". In: *Language* 98.1, pp. 1–34. URL: https://muse.jhu.edu/article/849525.

Beguš, G. (May 2023). "GPT4 draws a syntactic tree for the sentence: ..." Accessed on April 17, 2023. URL: https://twitter.com/begusgasper/status/1635719937275871233.

Beguš, G., M. Dabkowski, et al. (2025). "Large linguistic models: Investigating LLMs' metalinguistic abilities". Data deposit. OSF. URL: https://osf.io/y3bpt.

Beguš, G., A. Leban, et al. (2023). "Approaching an unknown communication system by latent space exploration and causal inference". In: *arXiv* 2303.10931. eprint: 2303.10931 (stat.ML).

Beguš, G. and A. Zhou (2022). "Modeling speech recognition and synthesis simultaneously: Encoding and decoding lexical and sublexical semantic information into speech with no direct access to speech data". In: *Proc. Interspeech 2022*, pp. 5298–5302. DOI: 10.21437/Interspeech.2022-11219.

Beguš, G., A. Zhou, and T. C. Zhao (2023). "Encoding of speech in convolutional layers and the brain stem based on language experience". In: *Scientific Reports* 13.1, p. 6480. DOI: 10.1038/s41598-023-33384-9. URL: https://doi.org/10.1038/s41598-023-33384-9.

Behzad, S. et al. (July 2023). "ELQA: A Corpus of Metalinguistic Questions and Answers about English". In: *Proceedings of the 61st Annual Meeting of the Association for Computational Linguistics (Volume 1: Long Papers)*. Toronto, Canada: Association for Computational Linguistics, pp. 2031–2047. DOI: 10.18653/v1/2023.acl-long.113. URL: https://aclanthology.org/2023.acl-long.113.

Bender, E. M., T. Gebru, et al. (2021). "On the dangers of stochastic parrots: Can language models be too big?" In: *Proceedings of the 2021 ACM conference on fairness, accountability, and transparency*, pp. 610–623.

Bender, E. M. and A. Koller (2020). "Climbing towards NLU: On meaning, form, and understanding in the age of data". In: *Proceedings of the 58th Annual Meeting of the Association for Computational Linguistics*, pp. 5185–5198.

Berglund, L. et al. (2024). *The Reversal Curse: LLMs trained on "A is B" fail to learn "B is A"*. arXiv: 2309.12288 [cs.CL]. URL: https://arxiv.org/abs/2309.12288.

Bialystok, E. et al. (1985). "Toward a Definition of Metalinguistic Skill". In: *Merrill-Palmer Quarterly* 31.3, pp. 229–251. ISSN: 0272930X, 15350266. URL: http://www.jstor.org/stable/23086295 (visited on 05/03/2023).

Blevins, J. (2004). *Evolutionary Phonology*. Cambridge: Cambridge University Press.

Blevins, T. et al. (July 2023). "Prompting Language Models for Linguistic Structure". In: *Proceedings of the 61st Annual Meeting of the Association for Computational Linguistics (Volume 1: Long Papers)*. Toronto, Canada: Association for Computational Linguistics, pp. 6649–6663. DOI: 10.18653/v1/2023.acl-long.367. URL: https://aclanthology.org/2023.acl-long.367.

Brown, T. B. et al. (2020). *Language Models are Few-Shot Learners*. arXiv: 2005.14165 [cs.CL].

Caucheteux, C. et al. (2022). "Brains and algorithms partially converge in natural language processing". In: *Communications biology* 5.1, p. 134.

Chalmers, D. J. (1993). "Connectionism and compositionality: Why Fodor and Pylyshyn were wrong". In: *Philosophical Psychology* 6.3.

Cho, H. et al. (2025). "Investigating ChatGPT's phonology problem-solving abilities through reasoning with varying custom instructions". In: *Linguistic Research* 42.1, pp. 53–93. DOI: 10.17250/khisli.42.1.202503.003.

Chomsky, N. (1993). "A Minimalist Program for Linguistic Theory". In: *The View from Building 20: Essays in Linguistics in Honor of Sylvain Bromberger*. Ed. by J. Keyser et al. MIT Press, pp. 1–52.

Chomsky, N. (1995). *The Minimalist Program*. Current Studies in Linguistics 28. Cambridge, MA: MIT Press, p. 420.

Chomsky, N. (2000). "Minimalist Inquiries: The Framework". In: *Step by step: Essays on minimalist syntax in honor of Howard Lasnik*. Ed. by R. Martin et al. Cambridge, MA: MIT Press, pp. 89–155.

Chomsky, N. (2014). "Minimal recursion: Exploring the prospects". In: *Recursion: Complexity in Cognition*. Studies in Theoretical Psycholinguistics 43. Springer, pp. 1–15.

Chomsky, N. and M. Halle (1968). *The Sound Pattern of English*. New York: Harper & Row.

Chomsky, N., R. A. Jacobs, et al. (1970). "Remarks on nominalization". In: *1970* 184, p. 221.

Cleeremans, A. (1993). *Mechanisms of implicit learning: Connectionist models of sequence processing*. MIT press.

Dahl, D. B. et al. (2019). xtable*: Export Tables to LaTeX or HTML*. R package version 1.8-4. URL: https://CRAN.R-project.org/package=xtable.

Davies, A. et al. (2021). "Advancing mathematics by guiding human intuition with AI". In: *Nature* 600.7887, pp. 70–74. DOI: 10.1038/s41586-021-04086-x. URL: https://doi.org/10.1038/s41586-021-04086-x.

Dentella, V. et al. (2023). "Systematic testing of three Language Models reveals low language accuracy, absence of response stability, and a yes-response bias". In: *Proceedings of the National Academy of Sciences* 120.51, e2309583120.

Dubey, A. et al. (2024). *The Llama 3 Herd of Models*. arXiv: 2407.21783 [cs.AI]. URL: https://arxiv.org/abs/2407.21783.

Everett, D. L. (2005). "Cultural constraints on grammar and cognition in Pirahã: Another look at the design features of human language". In: *Current Anthropology* 46.4, pp. 621–646.

Flavell, J. H. (1979). "Metacognition and cognitive monitoring: A new area of cognitive–developmental inquiry". In: *American Psychologist* 34.10, pp. 906–911. DOI: 10.1037/0003-066X.34.10.906.

Fodor, J. A. and B. P. McLaughlin (1990). "Connectionism and the problem of systematicity: Why Smolensky's solution doesn't work". In: *Cognition* 35.2, pp. 183–204.

Fodor, J. A. and Z. W. Pylyshyn (1988). "Connectionism and cognitive architecture: A critical analysis". In: *Cognition* 28.1, pp. 3–71. ISSN: 0010-0277. DOI: https://doi.org/10.1016/0010-0277(88)90031-5. URL: https://www.sciencedirect.com/science/article/pii/0010027788900315.

Goldstein, A. et al. (2022). "Shared computational principles for language processing in humans and deep language models". In: *Nature Neuroscience* 25.3, pp. 369–380.

Gulordava, K. et al. (June 2018). "Colorless Green Recurrent Networks Dream Hierarchically". In: *Proceedings of the 2018 Conference of the North American Chapter of the Association for Computational Linguistics: Human Language Technologies, Volume 1 (Long Papers)*. New Orleans, Louisiana: Association for Computational Linguistics, pp. 1195–1205. DOI: 10.18653/v1/N18-1108.

Haider, H. (2023). "Is Chat-GPT a grammatically competent informant?" Manuscript. Salzburg University. URL: https://lingbuzz.net/lingbuzz/007285.

Hauser, M. D. et al. (2002). "The faculty of language: What is it, who has it, and how did it evolve?" In: *Science* 298.5598, pp. 1569–1579.

Hockett, C. F. (1960). "The Origin of Speech". In: *Scientific American* 203.3, pp. 88–97.

Hu, J. et al. (2024). "Language models align with human judgments on key grammatical constructions". In: *Proceedings of the National Academy of Sciences* 121.36, e2400917121.

Jackendoff, R. et al. (2005). "The nature of the language faculty and its implications for evolution of language (Reply to Fitch, Hauser, and Chomsky)". In: *Cognition* 97.2.

Jumper, J. et al. (2021). "Highly accurate protein structure prediction with AlphaFold". In: *Nature* 596.7873, pp. 583–589. DOI: 10.1038/s41586-021-03819-2. URL: https://doi.org/10.1038/s41586-021-03819-2.

Kager, R. (1999). *Optimality Theory*. Cambridge Textbooks in Linguistics. Cambridge: Cambridge University Press.

Katzir, R. (2023). "Why large language models are poor theories of human linguistic cognition. A reply to Piantadosi (2023)". lingbuzz/007190. Manuscript. Tel Aviv University. URL: https://lingbuzz.net/lingbuzz/007190.

Kiparsky, P. (2006). "Amphichronic program vs. Evolutionary Phonology". In: *Theoretical Linguistics* 32.2, pp. 217–236.

Kodner, J. et al. (2023). "Why Linguistics Will Thrive in the 21st Century: A Reply to Piantadosi (2023)". In: *arXiv*. URL: https://arxiv.org/abs/2308.03228.

Lenth, R. (2018). *emmeans: Estimated Marginal Means, aka Least-Squares Means*. R package version 1.3.0. URL: https://CRAN.R-project.org/package=emmeans.

Lin, W. (2013). "Agnostic notes on regression adjustments to experimental data: Reexamining Freedman's critique". In: *The Annals of Applied Statistics* 7.1, pp. 295–318. DOI: 10.1214/12-AOAS583. URL: https://doi.org/10.1214/12-AOAS583.

Linzen, T. et al. (2016). "Assessing the Ability of LSTMs to Learn Syntax-Sensitive Dependencies". In: *Transactions of the Association for Computational Linguistics* 4, pp. 521–535. DOI: 10.1162/tacl_a_00115. URL: https://aclanthology.org/Q16-1037.

Matusevych, Y. et al. (2022). "*Trees neural those*: RNNs can learn the hierarchical structure of noun phrases". In: *Proceedings of the Annual Meeting of the Cognitive Science Society* 44.44, pp. 1848–1854.

Murty, S. et al. (2022). *Characterizing Intrinsic Compositionality in Transformers with Tree Projections*. arXiv: 2211.01288 [cs.CL].

OpenAI (2023). *GPT-4 Technical Report*. arXiv: 2303.08774 [cs.CL].

OpenAI (2024). "Learning to reason with LLMs". OpenAI Blog. URL: https://openai.com/index/learning-to-reason-with-llms.

Perruchet, P. et al. (2006). "Implicit learning and statistical learning: One phenomenon, two approaches". In: *Trends in cognitive sciences* 10.5, pp. 233–238.

Piantadosi, S. T. (2024). "Modern language models refute Chomsky's approach to language". In: *From fieldwork to linguistic theory: A tribute to Dan Everett*. Ed. by E. Gibson et al. Empirically Oriented Theoretical Morphology and Syntax 15. Berlin: Language Science Press, pp. 353–414. DOI: 10.5281/zenodo.12665933. URL: https://doi.org/10.5281/zenodo.12665933.

Prince, A. et al. (1993). "Optimality Theory: Constraint Interaction in Generative Grammar". In: *Rutgers Optimality Archive*. ROA-537, ROA Version, 8/2002. URL: https://roa.rutgers.edu/article/view/547.

R Core Team (2024). *R: A Language and Environment for Statistical Computing*. R Foundation for Statistical Computing. Vienna, Austria. URL: https://www.R-project.org/.

Reber, A. S. (1967). "Implicit learning of artificial grammars". In: *Journal of verbal learning and verbal behavior* 6.6, pp. 855–863.

Renze, M. et al. (2024). *The Effect of Sampling Temperature on Problem Solving in Large Language Models*. DOI: 10.48550/arXiv.2402.05201. arXiv: 2402.05201 [cs.CL]. URL: https://arxiv.org/abs/2402.05201.

Şahin, G. G. et al. (2020). "PuzzLing Machines: A Challenge on Learning from Small Data". In: *Proceedings of the 58th Annual Meeting of the Association for Computational Linguistics*. Association for Computational Linguistics, pp. 1241–1254. URL: https://aclanthology.org/2020.acl-main.115.pdf.

Schaeffer, R. et al. (2023). *Are Emergent Abilities of Large Language Models a Mirage?* arXiv: 2304.15004 [cs.AI].

Smolensky, P. (1988). "On the proper treatment of connectionism". In: *Behavioral and Brain Sciences* 11.1, pp. 1–23.

Tunmer, W. E. et al. (1984). "The Development of Metalinguistic Awareness: A Conceptual Overview". In: *Metalinguistic Awareness in Children: Theory, Research, and Implications*. Ed. by W. E. Tunmer et al. Berlin, Heidelberg: Springer Berlin Heidelberg, pp. 12–35. ISBN: 978-3-642-69113-3. DOI: 10.1007/978-3-642-69113-3_2. URL: https://doi.org/10.1007/978-3-642-69113-3_2.

Vaswani, A. et al. (2017). "Attention is All you Need". In: *Advances in Neural Information Processing Systems*. Ed. by I. Guyon et al. Vol. 30. Curran Associates, Inc.

Vaux, B. (2008). "Why the Phonological Component Must Be Serial and Rule-Based". In: *Rules, Constraints, and Phonological Phenomena*. Ed. by B. Vaux et al. Oxford: Oxford University Press, pp. 20–61.

Wilcox, E. et al. (Nov. 2018). "What do RNN Language Models Learn about Filler–Gap Dependencies?" In: *Proceedings of the 2018 EMNLP Workshop BlackboxNLP: Analyzing and Interpreting Neural Networks for NLP*. Brussels, Belgium: Association for Computational Linguistics, pp. 211–221. DOI: 10.18653/v1/W18-5423.

Yedetore, A. et al. (2023). "How poor is the stimulus? Evaluating hierarchical generalization in neural networks trained on child-directed speech". In: *arXiv preprint arXiv:2301.11462*.

Živanović, S. (2017). *Forest*. https://ctan.org/pkg/forest. Version 2.1.5. LaTeX package for drawing (linguistic) trees.

## Acknowledgments

This research was partially funded by the Georgia Lee Fellowship in the Society of Hellman Fellows as well as a grant by Project CETI to Gašper Beguš.

We would like to thank Alexander Elias, Allegra Robertson Molinaro, and Kai Joseph Schenck for their contributions to task evaluation. Gašper Beguš would also like to thank students in his *Linguistics 111: Phonology* and *Linguistics 211A: Advanced Phonology I* classes at the University of California, Berkeley.